\documentclass[journal]{IEEEtran}

\usepackage{graphics}
\usepackage{pdfpages}
\usepackage{cite}
\usepackage{amsmath}
\usepackage{amsfonts}
\usepackage{bm}
\usepackage{comment}
\usepackage{booktabs}
\usepackage{multirow}
\usepackage{mathtools}
\usepackage{threeparttable}
\usepackage{siunitx}
\usepackage{amssymb}
\usepackage{pifont}

\usepackage{algorithmicx}
\usepackage[ruled]{algorithm}
\usepackage[noend]{algpseudocode}

\newcommand{\figlab}[1]{\label{fig:#1}}
\newcommand{\figref}[1]{Fig.~\ref{fig:#1}} 
\newcommand{\tablab}[1]{\label{tab:#1}}
\newcommand{\tabref}[1]{Table~\ref{tab:#1}} 
\newcommand{\forlab}[1]{\label{for:#1}}
\newcommand{\forref}[1]{Equation~(\ref{for:#1})} 

\newcommand{\algolab}[1]{\label{algorithm:#1}}
\newcommand{\algoref}[1]{Algorithm~\ref{algorithm:#1}} 

\newcommand{\ie}{\textit{i}.\textit{e}.}
\newcommand{\eg}{\textit{e}.\textit{g}.}
\newcommand{\etal}{\textit{et al}.}

\usepackage{url}
\usepackage{subcaption}

\begin{document}

\title{Many-Objective-Optimized Semi-Automated Robotic Disassembly Sequences}

\author{Takuya~Kiyokawa,~\IEEEmembership{Member,~IEEE}, Kensuke~Harada,~\IEEEmembership{Senior Member,~IEEE}, Weiwei~Wan,~\IEEEmembership{Senior Member,~IEEE}, Tomoki~Ishikura, Naoya~Miyaji, and Genichiro~Matsuda
\thanks{T. Kiyokawa, K. Harada, and W. Wan are with Department of Systems Innovation, Graduate School of Engineering Science, Osaka University, 1-3 Machikaneyama-cho, Toyonaka-shi, Osaka, Japan (email: {kiyokawa, harada, wan}@sys.es.osaka-u.ac.jp)}
\thanks{K. Harada is with the Industrial Cyber-physical Systems Research Center, The National Institute of Advanced Industrial Science and Technology (AIST), 2-3-26 Aomi, Koto-ku, Tokyo, Japan (email: kensuke.harada@aist.go.jp)}
\thanks{T. Ishikura, N. Miyaji, and G. Matsuda are with Manufacturing Innovation Division, Panasonic Holdings Corporation, 2-7 Matsuba-cho, Kadoma, Osaka, Japan (email: {ishikura.tomoki, miyaji.naoya, matsuda.genichiro}@jp.panasonic.com)}}

\markboth{}%
{Kiyokawa \MakeLowercase{\etal}: Many-Objective-Optimized Semi-Automated Robotic Disassembly Sequences}

\maketitle

\begin{abstract} 
This study tasckles the problem of many-objective sequence optimization for semi-automated robotic disassembly operations. To this end, we employ a many-objective genetic algorithm (MaOGA) algorithm inspired by the Non-dominated Sorting Genetic Algorithm (NSGA)-III, along with robotic-disassembly-oriented constraints and objective functions derived from geometrical and robot simulations using 3-dimensional (3D) geometrical information stored in a 3D Computer-Aided Design (CAD) model of the target product. The MaOGA begins by generating a set of initial chromosomes based on a contact and connection graph (CCG), rather than random chromosomes, to avoid falling into a local minimum and yield repeatable convergence. The optimization imposes constraints on feasibility and stability as well as objective functions regarding difficulty, efficiency, prioritization, and allocability to generate a sequence that satisfies many preferred conditions under mandatory requirements for semi-automated robotic disassembly.
The NSGA-III-inspired MaOGA also utilizes non-dominated sorting and niching with reference lines to further encourage steady and stable exploration and uniformly lower the overall evaluation values. Our sequence generation experiments for a complex product (36 parts) demonstrated that the proposed method can consistently produce feasible and stable sequences with a 100\% success rate, bringing the multiple preferred conditions closer to the optimal solution required for semi-automated robotic disassembly operations.
\end{abstract}


\section{Introduction}
\IEEEPARstart{W}{ith} the aim of achieving a sustainable society, there has been greater emphasis on promoting recycling, reuse, and remanufacturing. In particular, future manufacturing robots are expected to exhibit proficiency in efficient disassembly of many parts. In this context, autonomous robotic disassembly has garnered increased attention~\cite{Poschmann2020,Daneshmand2022}. 
The remanufacturing domain requires the deployment of robots capable of autonomously acquiring sequential disassembly operations in an efficient and streamlined manner to address a diverse array of needs. Furthermore, human-robot cooperation has been promising for carefully extracting valuable parts from disassembly target products~\cite{Lou2023}.

To achieve the sequential disassembly operations without much manual effort, the automatic generation of sequences is crucial. To automatically generate sequences for (dis)assembly, previous studies have employed a three-dimensional (3D) model of the product~\cite{Thomas2018,Tariki2021,Chervinskii2021,Koga2022}. Determining the order of (dis)assembly parts can be categorized as a combinatorial optimization problem and a Non-deterministic Polynomial-time (NP)-hard problem~\cite{Goldwasser1996}, which necessitates the use of heuristic search algorithms to obtain a suboptimal solution within a practical timeframe.
\begin{figure}[tb]
  \centering
  \includegraphics[width=\linewidth]{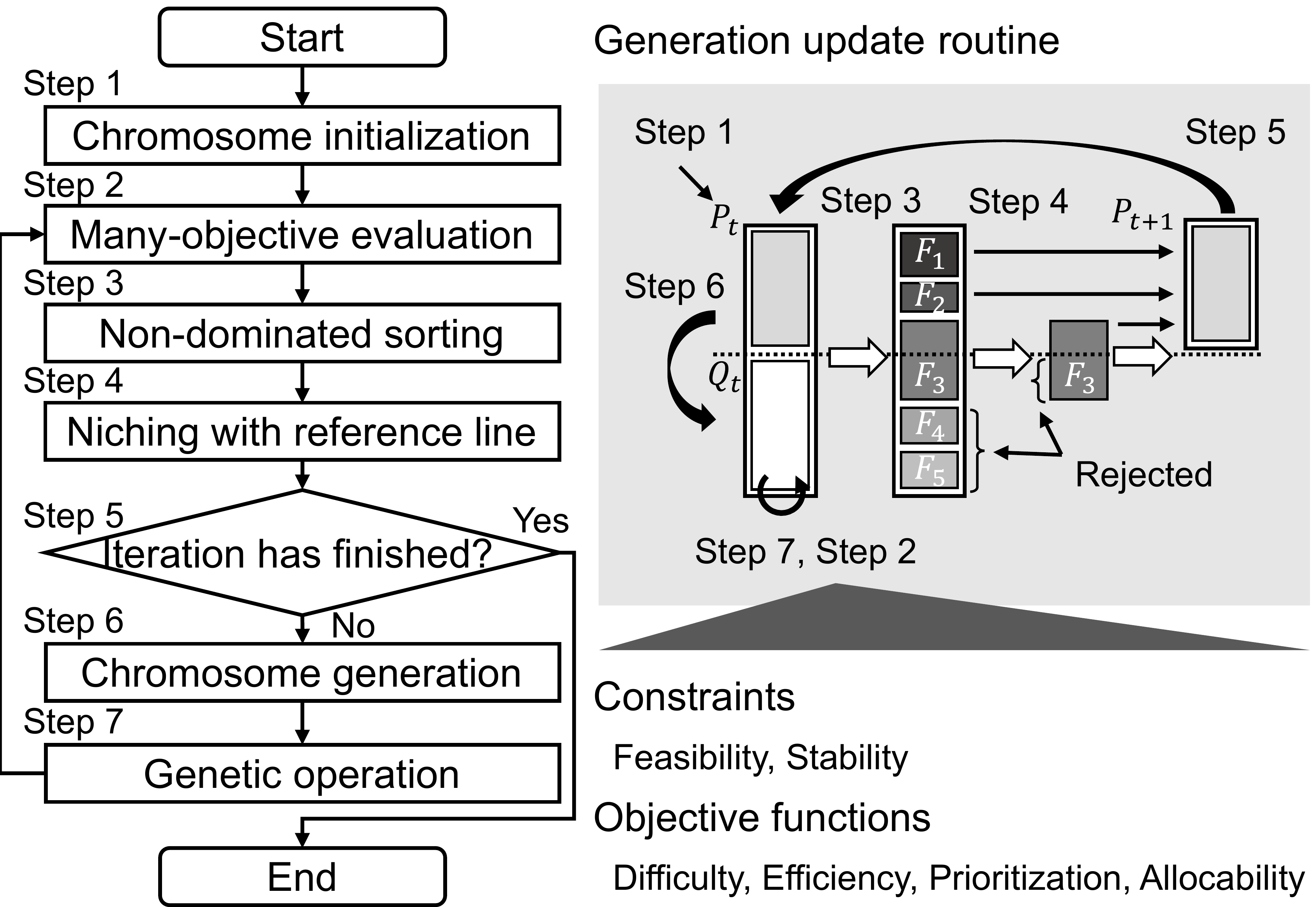}
  \caption{\small{NSGA-III-inspired sequence optimization.}}
  \figlab{nsga3}
\end{figure}
\begin{figure}[tb]
  \begin{minipage}[tb]{\linewidth}
      \centering
      \includegraphics[width=0.94\linewidth]{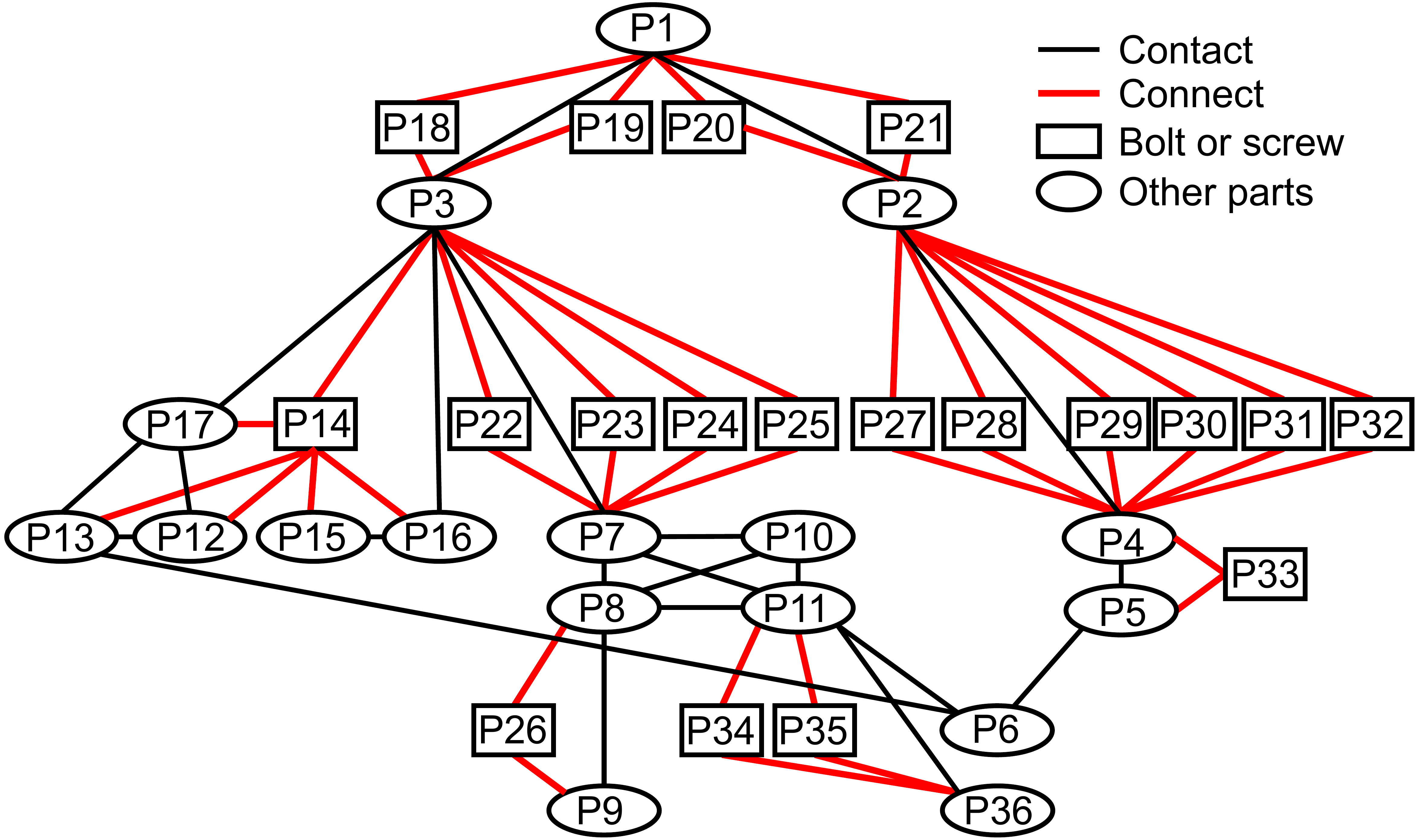}
      \subcaption{\small{Contact and connection graph (CCG).}}
  \end{minipage}
  \begin{minipage}[tb]{\linewidth}
      \centering
      \includegraphics[width=0.932\linewidth]{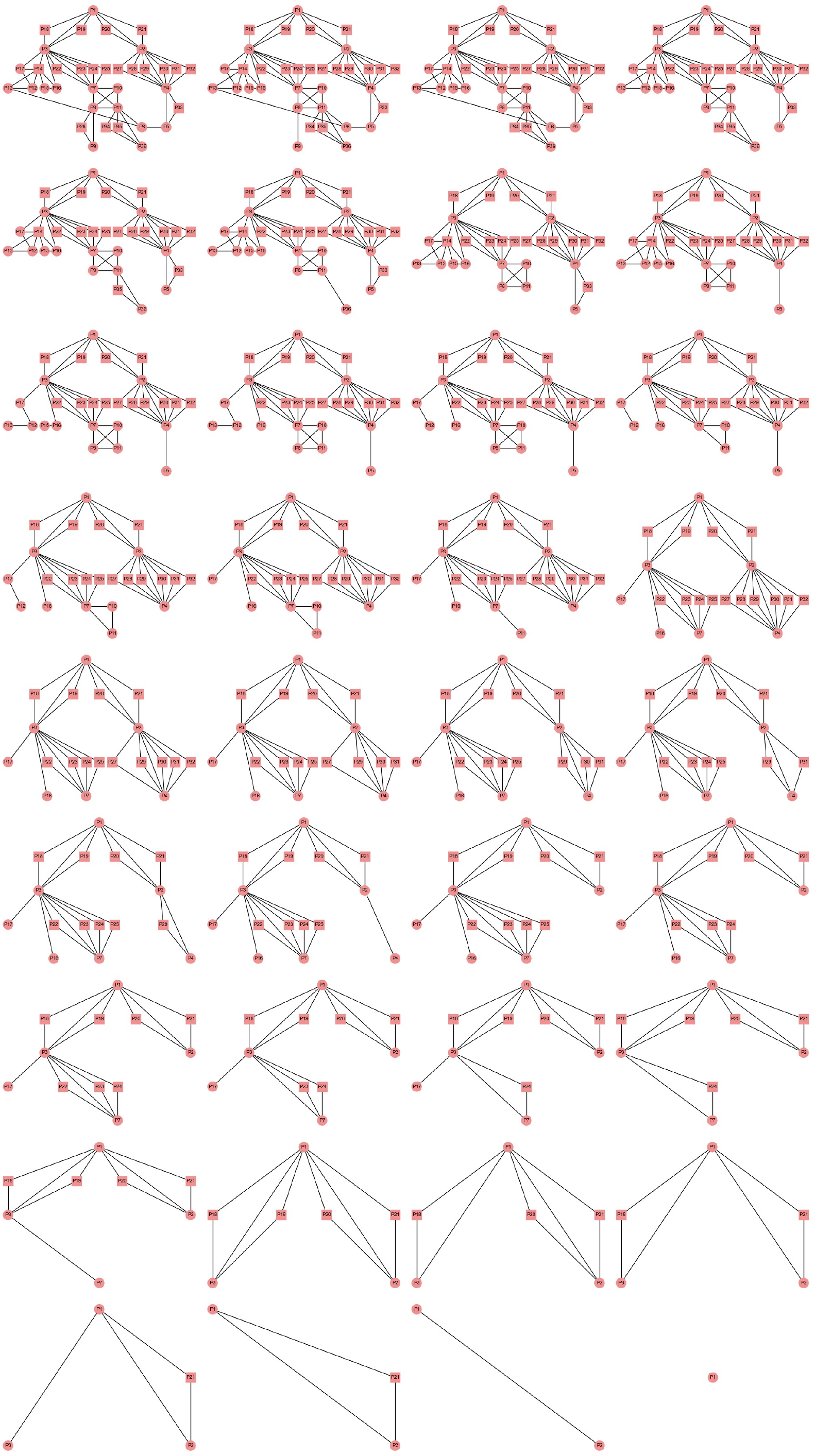}
      \subcaption{\small{An example of the generated initial solution.}}
  \end{minipage}
  \caption{\small{CCG-based initialization (CCGI).}}
  \vspace{-2mm}
  \figlab{graph}
\end{figure}

To facilitate robotic disassembly for flexible remanufacturing effectively, it is crucial to consider various aspects when determining the required sequence.
Several studies have explored disassembly sequence planning (DSP) in the context of semi-automated processes or human-robot-collaboration (HRC) by considering different perspectives~\cite{Chen2014,Parsa2021,Liao2022,Lou2023,Hjorth2023,Lee2023,Guo2023,Lee2023}.
To generate a disassembly sequence, they evaluated the disassembly cost, disassemblability, and safety from the long-horizon perspective of the sequence to optimize the overall order of parts and processes in real-time.
However, these studies did not address MaOPs under constraints and objective functions specific to semi-automated robotic disassembly with the planning of robot operations, as in our study.
Therefore, this study addresses robotic DSP rather than merely optimizing the preferable parts order and desirable processes.

Specifically, this study establishes constraints on disassembly order feasibility, robot motion feasibility, and object placement stability.
The designed objective functions pertain to order difficulty based on the contact state transition difficulty, task efficiency determined by the number of end-effector (eef) changes and distance between adjacent pairs in the sequence, prioritization based on the user-defined priority of disassembling particular parts before others, and allocability based on the number of task-allocated agent changes between humans and robots.

To address the many-objective optimization problem (MaOP) with conflicting objectives, we employ a Many-Objective Genetic Algorithm (MaOGA) inspired by the Non-Dominated Sorting Genetic Algorithm III (NSGA-III)~\cite{NSGAIII}, a state-of-the-art evolutionaly many-objective optimization (EMaO) algorithm.
\figref{nsga3} shows an overview of the proposed algorithm.
Generally, to avoid convergence to local optimal solutions, heuristic search algorithms need to effectively generate as many constraint-satisfied initial solutions as possible.
To generate chromosomes represented as a set of sequences, we propose to generate a contact and connection graph (CCG) representing the contact and connection relationships among disassembly parts and use the CCG to efficiently generate constraint-satisfied initial solutions.
The generation of the disassembly order of parts can be achieved through simple step-by-step removal of the end nodes of the graph.
\figref{graph}~(a) shows an example of the CCG for the target product used for our experiments in this study.

This study examines the effectiveness of the proposed stability-based initial chromosome generation method depicted in~\figref{graph}~(b) by verifying that it outperforms random chromosome initialization.
The NSGA-III-inspired algorithm also utilizes non-dominated sorting and niching with reference lines to further encourage steady and stable exploration of the solutions and uniformly lower the overall evaluation values.
Our simulation experiment verify the applicability of the newly-constructed NSGA-III-inspired algorithm tailored for the robotic DSP problem.

In the evaluation, we show that chromosome initialization repeatedly generates the most interference-free and stable initial solutions by comparing with other initialization methods.
Our ablation study further show that the NSGA-III-inspired algorithm can steadily and effectively reduce the evaluation values through many-objective optimization based on non-dominated sorting and niching with reference lines.
Finally, the proposed algorithm successfully generated disassembly sequences and robotic disassembly operations for a complex belt drive unit composed of 36 parts while considering multiple necessary constraints and desirable objectives from diverse perspectives, including semi-automated robotic operations.

In the rest of this paper, we discuss related work in Section II, describe our proposed algorithm in Section III, present the evaluation results in Section IV, discuss the remaining issues in Section V, and conclude the paper in Section VI.

\section{Related Work}
\subsection{Determining Order of Parts}
The optimization and determination of the order of parts in (dis)assembly processes have been the focus of extensive research for several decades~\cite{Mello1988,Lee1999,Sundaram2001,Le2009,Tariki2021,Zhao2021,Kiyokawa2021seq,Laili2022,Dorn2022,Wang2022,Liao2022,Tian2022,Lee2023}. The growing demand for high-mix, low-volume production and significant advancements in computer capabilities have recently revived interest in this topic in the field.

Kiyokawa~\etal~\cite{Kiyokawa2021seq} proposed an multi-objective genetic algorithm (MOGA) specifically tailored for assembly sequence planning (ASP).
The experimental outcome suggest that the proposed NSGA-II~\cite{NSGAII}-inspired MOGA can identify Pareto optimal solutions across a range of objective functions that assess the interference, insertion, and constraint relationships between parts. Our study employs NSGA-III, which is an extension of this ASP-specific NSGA-II algorithm.

Ebinger~\etal~\cite{Ebinger2018} introduced a flexible DSP framework which includes a subassembly identification method. The authors investigated the usefulness of subassemblies in search by examining the framework performance with and without subassemblies.
Chervinskii~\etal~\cite{Chervinskii2021} introduced Auto-Assembly, a framework that encompasses design analysis, ASP, Bill-of-Process (BOP) generation, and control code execution for physical assembly.
Dorn~\etal~\cite{Dorn2022} addressed the challenge of generating a Voronoi diagram for a complex automotive model, and developed an assembly priority graph.
Wang~\etal~\cite{Wang2022} focused on robotic parallel DSP for end-of-life products and proposed a multi-objective model to minimize the makespan and energy consumption.

Several studies have solved several cases of path-planning problems for ASP and DSP.
Lee~\etal~\cite{Lee2000} achieved minimization of the steps to goal component removal based on the concept of blocking topology.
Le~\etal~\cite{Le2009} extended a sampling-based path planner Rapidly-exploring Random Tree (RRT) for ASP and DSP for LEGO bricks.
Moreover, Tian~\etal~\cite{Tian2022} proposed a physics-based assembly-by-disassembly planner using a large-scale dataset, achieving state-of-the-art performance.

However, these studies~\cite{Kiyokawa2021seq,Ebinger2018,Chervinskii2021,Dorn2022,Wang2022,Lee2000,Le2009,Tian2022} did not address the finer aspects of robot motions, such as grasps and trajectories, which are critical for robotic disassembly operations.

Recent studies have employed graph neural networks (GNNs) to deduce a feasible sequence by analyzing the graph representation of the complex (dis)assembly model structure. 
Cebulla~\etal~\cite{Cebulla2023} introduced an assembly-by-disassembly approach that involves iteratively testing parts for removal, with the testing order significantly impacting runtime.
The authors optimized the order using a GNN trained on part shapes and local connections.
Ma~\etal~\cite{Ma2023} introduced a heterogeneous graph-transformer framework for learning the latent rules of assembly planning.
However, these studies~\cite{Cebulla2023,Ma2023} did not address robotic disassembly and limited the assembly targets to aluminum profiles or LEGO bricks consisting of a small number of parts compared to actual mechanical products.
Recent studies on multi-objective optimization of disassembly orders~\cite{Parsa2021,Ren2021,Zhao2021,Guo2023} have focused on improving heuristic algorithms, rather than establishing an optimization framework that considers robotic disassembly operations.

\subsection{Robotic Assembly and Disassembly}
On the other hand, the process of planning (dis)assembly robot motions necessitates taking into account constraints in multiple dimensions and perspectives.
Thus far, several researchers have successfully generated and executed disassembly operations using robots, assuming that the (dis)assembly order is given or that the operations involve two separate parts~\cite{Dario1994,Hohm2000,Zussman2000,Kim2013}.

Rodriguez~\etal~\cite{Rodriguez2019} proposed iteratively checking multilevel feasibilities to plan assembly sequences with robot motions.
Bachmann~\etal~\cite{Bachmann2021} investigated the impact of robotic workcell layout on task efficiency and feasibility.
Most recently, Atad~\etal~\cite{Atad2023} used a GNN to infer feasible and optimal assembly sequences by learning an inference model based on a geometric structure graph representation of a product. 
However, these studies~\cite{Rodriguez2019,Bachmann2021,Atad2023} more focused on the aluminum profile assemblies consisting of a small number of parts compared to mechanical products.

Liu~\etal~\cite{Liu2020} succeeded in optimizing the sequence and process of robotic disassembly through collaborative optimization; however, they did not take into account feasible motions (\ie, contacts and trajectories).
Koga~\etal~\cite{Koga2022} proposed a Computer-Aided Design (CAD)-based robotic assembly system; however, generating sequential and semi-automated disassembly operations optimized using multiple objective functions remains an open issue.

Laili~\etal~\cite{Laili2022} addressed the issue of flexible sequencing in robotic disassembly with failed automation operations, proposing an online recovery method that utilizes pre-stored backup actions.
Jiayi~\etal~\cite{Jiayi2023} used a digital twin for dynamic planning of robotic disassembly under uncertain real-world conditions. 
These studies~\cite{Laili2022,Jiayi2023} focused on different issues from our study.
Learning-based assembly planning approaches~\cite{Thomas2018,Zakka2020,Lee2021,Aslan2022,Qu2023} have demonstrated promising results in adapting to a broader range of products; however, their efficacy has primarily been validated with toy objects or furniture.
Unlike these studies, this study does not involve the exploration of dynamic planning and learning-based generalization.

This study does not delve into other issues related to disassembly specific system components, including eefs, controllers, and interfaces, as reported in~\cite{Chen2014,Borras2018,Klas2021,Wurster2022,Zhang2023,Hjorth2023}.
Gorjup~\etal~\cite{Gorjup2020} introduced an integrated flexible manufacturing system that uses compliance control, CAD-based localization, and multimodal gripper for fast and efficient assembly task programming. However, this system was limited to part-kitting tasks only.

In contrast, our study aims to explore the feasibility of generating effective disassembly sequences for a complex product by designning suitable constraints, objective functions, chromosome initialization rules, and genetic generation update rules in the proposed EaMO method.

\section{Optimizing Robotic Disassembly Sequence}
\alglanguage{pseudocode}
\begin{algorithm}[t]
\caption{Robotic DSP} \algolab{dsp}
\begin{algorithmic}[1]
\renewcommand{\algorithmicrequire}{\textbf{Input:}}
\renewcommand{\algorithmicensure}{\textbf{Output:}}
\Require 3D CAD models of parts and environment $\bm{M}$
\Ensure An optimum sequence $\bm{\hat{O}}$, disassembly task label $\bm{\mathcal{\hat{L}}}$, eef contact $\bm{\mathcal{\hat{C}}}$, arm trajectory $\bm{\mathcal{\hat{T}}}$, and object pose $\bm{\mathcal{\hat{P}}}$
\State $\bm{\mathcal{I}}$, $\bm{X}^{if}, \bm{X}^{cs}, \bm{X}^{ct}, \bm{X}^{cf}$ = GeometricalSimulation($\bm{M}$)
\State $\bm{X}^{mf}, \bm{\mathcal{C}}, \bm{\mathcal{T}}, \bm{\mathcal{P}}$ = RobotSimulation($\bm{M}$)
\State Set $T^i_\mathrm{max}, T^g_\mathrm{max}, \bm{P}^{ga}$ with user inputs
\State SetPlanner($\bm{\mathcal{I}}, \bm{X}^{if}, \bm{X}^{cs}, \bm{X}^{ct}, \bm{X}^{cf}, \bm{X}^{mf}, \bm{\mathcal{C}}, \bm{\mathcal{T}}, \bm{\mathcal{P}}, \bm{P}^{ga}$) 
\State Initialize the counters of $t^i$ and $t^g$ to 1
\While {$t^i < T^i_\mathrm{max}$}
    \State $\bm{G}^1 =$ ChromosomeInitialization()
    \While {true}
        \State Initialize the elements of $\bm{E}^{G^{t^g}}$ to 1
        \State $\bm{C}^{G^{t^g}}$ = ConstraintCheck($\bm{G}^{t^g}$)
        \If {All elements of $\bm{C}^{G^{t^g}}$ are \textit{Available}}
            \State $\bm{E}^{G^{t^g}}$ = FitnessCalculation($\bm{G}^{t^g}$)
        \EndIf
        \State $\bm{G}^{best} =$ BestSolutionExtraction($\bm{G}^{t^g}$, $\bm{E}^{G^{t^g}}$)
        \If {$T^{g}_\mathrm{max} < t^g$}
            \State \textbf{break}
        \EndIf
        \State $\bm{\tilde{G}}^{t^g}$, $\bm{E}^{\tilde{G}^{t^g}}$ = NonDominatedSorting($\bm{G}^{t^g}$, $\bm{E}^{G^{t^g}}$)
        \State NichingWithReferenceLine($\bm{\tilde{G}}^{t^g}$, $\bm{E}^{\tilde{G}^{t^g}}$)
        \State $\bm{G}^{t^g+1}$ = NextGenerationCreation($\bm{\tilde{G}}^{t^g}$, $\bm{E}^{\tilde{G}^{t^g}}$)
        \State $\bm{G}^{t^g+1}$ = GeneticOperation($\bm{G}^{t^g+1}$)
        \State $t^g$ = $t^g$ + 1
    \EndWhile
    \State $t^i$ = $t^i$ + 1
\EndWhile
\State $\bm{\hat{O}} = \bm{G}^{best}$
\State $\bm{\mathcal{\hat{L}}} =$ LabelExtraction($\bm{\hat{O}}$)
\State $\bm{\mathcal{\hat{C}}}$, $\bm{\mathcal{\hat{T}}}$, $\bm{\mathcal{\hat{P}}}$ $=$ OperationParameterExtraction($\bm{\hat{O}}$)
\end{algorithmic}
\end{algorithm}
\subsection{Overview}
\algoref{dsp} outlines the flow of the robotic DSP algorithm.
In our algorithm, the order is represented by $\bm{O}_k~(k=1,\ldots,N^p)$, where ${N^p}$ represents the number of parts.
The last part is denoted by $\bm{O}_1$, and the first part is denoted by $\bm{O}_{N^p}$.
The order obtained through optimization is denoted by $\bm{\hat{O}}$.
The algorithm takes 3D CAD models of the target parts and environment (\eg, a robot arm, gripper, and worktable), represented by $\bm{M}$, as inputs.
The outputs of the algorithm include the optimized sequence $\bm{\hat{O}}$, along with the task labels $\bm{\mathcal{\hat{L}}}$, eef contacts $\bm{\mathcal{\hat{C}}}$, arm trajectories $\bm{\mathcal{\hat{T}}}$, and object placement poses $\bm{\mathcal{\hat{P}}}$.

Our DSP process commences with an in-depth analysis of the geometries derived from CAD models $\bm{M}$, subsequently generating parts-relation matrices.
The first step involves extracting the labels associated with part names, as well as information pertaining the center of mass, pose, and shape of each part, represented by $\bm{\mathcal{I}}$.
Subsequently, the GeometricalSimulation($\bm{M}$) function provides us with interference-free $\bm{X}^{if}$, constraint degree $\bm{X}^{cs}$, contact $\bm{X}^{ct}$, and constraint-free $\bm{X}^{cf}$ matrices, which encapsulate diverse types of relationships between each pair of parts.

The function RobotSimulation($\bm{M}$) provides the motion-feasibility matrices $\bm{X}^{mf}_i~(i=1,\ldots,N^p)$, which contain the feasible contacts $\bm{\mathcal{C}}$, trajectories $\bm{\mathcal{T}}$, and placement poses $\bm{\mathcal{P}}$ for each part.
After planning the contacts to initiate the manipulation of the target part $P_i$ and generating a multitude of collision-free contact samples, each sample is evaluated for collision-free and Inverse Kinematics (IK)-solvable trajectories for all specified placement poses. 
Furthermore, we check if there exists a trajectory enabling the undisassembled target part to be moved from the placement pose to other placement poses (possible next placement pose) by grasping and relocating it with a two-fingered hand.
The dataset used in this study consists of all possible combinations of placement poses for all possible compositions of the undisassembled target object, which has been previously examined. The dataset is saved in graph format, as described in a previous study~\cite{Wan2017}.
For both the disassembly trajectory and transportation trajectory between the two placement poses, if at least one trajectory set is generated, the corresponding binary element of $\bm{X}^{mf}_i$ can be 1 (feasible).
If multiple feasible trajectory sets exist for the same placement poses, the shortest trajectory is selected to guarantee minimal trajectory length.
The feasibility is stored in the corresponding element of $\bm{X}^{mf}_i$.
The elements of $\bm{X}^{mf}_i$ indicate whether each feasible motion (including feasible grasp, trajectories, and object placement) generated for the target part $P_{O_i}$ interferes with parts other than $P_{O_i}$.

The user specifies the MaOGA parameters, including the maximum number of iterations $T^i_\mathrm{max}$,the maximum number of generation updates $T^g_\mathrm{max}$, and a set of other MaOGA parameters $\bm{P}^{ga}$.
$\bm{P}^{ga}$ includes the number of chromosomes, crossover rate, mutation rate, cut-and-paste rate, and break-and-join rate.
The matrices and parameters are then set to the planner using the SetPlanner() function at line~4~in~\algoref{dsp}.
Once the planner is set, the matrices and parameters can be accessed from any function in the algorithm.

Prior to initiating the optimization loop, the values of $t^i$ and $t^g$ are both set to 1.
As illustrated in~\figref{nsga3}, optimization process commences with the initialization of the first-generation chromosome $\bm{G}^1$ using ChromosomeInitialization(). Subsequently, the optimization (generation update) loop commences, where genes are evaluated using ConstraintCheck($\bm{G}^{t^g}$) and FitnessCalculation($\bm{G}^{t^g}$). The best solution $\bm{G}^{best}$ is extracted using with BestSolutionExtraction($\bm{G}^{t^g}$, $\bm{E}^{G^{t^g}}$).
Additionally, based on the evaluation values $\bm{E}^{G^{t^g}}$, $\bm{G}^{t^g}$ is sorted using NonDominatedSorting($\bm{G}^{t^g}$, $\bm{E}^{G^{t^g}}$).
The sorted solutions $\bm{\tilde{G}}^{t^g}$ along with their evaluation values~$\bm{E}^{\tilde{G}^{t^g}}$ are associated with the reference lines using the NichingWithReferenceLine($\bm{\tilde{G}}^{t^g}$, $\bm{E}^{\tilde{G}^{t^g}}$).
Reference lines are created from points at infinity to the optimal point.
The assignment of solutions with reference lines guides the exploration direction of each solution.
By repeatedly applying this process, NSGA-III steers the exploration.
Consequently, when the algorithm converges, the reference lines facilitate the discovery of a superior representative non-dominated solution.

Following the selection of solutions (sequences) to which the genetic operations are applied based on the assignment with the reference line, the process advances to the creation of the next generation $\bm{G}^{t^g+1}$ through the NextGenerationCreation($\bm{\tilde{G}}^{t^g}$, $\bm{E}^{\tilde{G}^{t^g}}$) function.
The genetic information of the selected genes is altered through the application of genetic operations with the GeneticOperation($\bm{G}^{t^g+1}$) function.
In this study, we used the four genetic operators proposed in~\cite{Tariki2021}, namely crossover, mutation, cut-and-paste, and break-and-join.
The process is repeated until the values of $t^i$ and $t^g$ reach $T^i_\mathrm{max}$ and $T^g_\mathrm{max}$, respectively.

Upon completion, the optimal solution $\bm{G}^{best}$ is stored in $\bm{\hat{O}}$.
The disassembly task labels $\bm{\mathcal{\hat{L}}}$ are extracted from the datasrts obtained from the input CAD model, using LabelExtraction($\bm{\hat{O}}$), at the commencement of the algorithm.
The eef contacts $\bm{\mathcal{\hat{C}}}$, arm trajectories $\bm{\mathcal{\hat{T}}}$, and object placement poses $\bm{\mathcal{\hat{P}}}$ are extracted from the datasets generated through the robot simulations conducted at the beginning of the algorithm, utilizing OperationParameterExtraction($\bm{\hat{O}}$).

In the evaluation process, the feasibility and stability of each solution (sequence) $\bm{G}^{t^g}_i~(i=1,\ldots,N^{gn})$ are determined using the aforementioned matrices, resulting in constraint satisfiability $\bm{C}^{G^{t^g}}$ consisting of binary elements that indicate whether each solution (sequence) is \textit{feasible}, \textit{stable}, and \textit{available}.
The number of genes is denoted by $N^{gn}$.
A solution is considered available if it is both feasible and stable, \ie, available $\coloneqq$ feasible $\land$ stable. 
The available solutions (sequences) are further sorted based on multiple criteria such as difficulty, efficiency, prioritization, and allocability.
The evaluation provides evaluation values $\bm{E}^{G^{t^g}}$.
The proposed MaOGA method resolves the minimization problem by employing multiple objective functions that are normalized between 0 and 1, with 0 representing the optimal value.

\subsection{CAD-Informed Matrices Generations}
This section outlines the preprocessing steps before the optimization loop, including the definition of part labels and the structure analysis of 3D CAD models. Matrix generation is also discussed, with a focus on determining constraints and calculating evaluation values.
Part labels are assigned to each component of the model, as listed in~\tabref{spec}. These labels include task labels, which link the parts to the eefs of the robot; a priority label, which designates parts that should be disassembled preferentially; a base label, which is assigned to the root part fixed at the end of the sequence; and an ignore label, which is applied to parts that can be disregarded in the DSP as they are automatically disassembled during the removal of other parts.

The task labels are designated to specific parts, including screws, bolts, nuts, plates, graspable objects, and manually disassembled objects.
The graspable label is assigned to parts other than screws, bolts, nuts, and plates, and can be manipulated by a two-finger gripper.
The manual label is assigned to parts that are difficult to automate or necessitate special care when disassembling them. The priority and value labels are established for parts that merit prioritization during disassembly and are classified as valuable in terms of reuse, recycling, and remanufacturing.
\begin{table}[tb]
    \centering
        \caption{Required labels for parts} \tablab{spec}
        \begin{tabular}{lll} \toprule
            \begin{tabular}{c}Label\end{tabular}
            &\begin{tabular}{c}Class\end{tabular} \\ \midrule
            Task & screw, bolt, nut, plate, graspable, manual \\
            Priority & value \\
            Base & base \\
            Ignore & ignore \\ \bottomrule
        \end{tabular}
\end{table}

To analyze the 3D CAD models (\eg, label extraction), our framework utilizes PythonOCC\footnote{https://dev.opencascade.org/project/pythonocc}, which is a Python wrapper for the OpenCASCADE\footnote{https://www.opencascade.com/} library.
PythonOCC converts the Standard for the Exchange of Product (STEP) model data into a format that allows for the extraction of part names, 3D poses, and shapes of the target product.
The label for each part is obtained by splitting the part name using an underscore as the delimiter.

To obtain a quantifiable representation of part relationships, we employed previously proposed matrix representations~\cite{Tariki2021,Kiyokawa2021seq}.
Tariki~\etal~\cite{Tariki2021} utilized PythonOCC to generate the interference-free matrices~$\bm{X}^{if}_j~(j=1,\ldots,6)$ with elements indicating binary values of interference or interference-free between each set of two parts.
Kiyokawa~\etal~\cite{Kiyokawa2021seq} presented a method for generating a constraint degree matrix~$\bm{X}^{cs}$ and contact matrix~$\bm{X}^{ct}$, which indicates the degree of constraint from 0 to 12 and the binary values of contact between each set of two parts.
In this study, we develop constraint-free matrices~$\bm{X}^{cf}_j~(j=1,\ldots,12)$ and the motion feasibility matrices~$\bm{X}^{mf}_i~(i=1,\ldots,N^p)$.
As the total number of elements in the interference-free matrix, constraint degree matrix, contact matrix, constrant-free matrix, and motion feasibility matrix is $20\times{N^p}\times{N^p}+\sum_{i=1}^{N^p} N^m_i\times{N^p} $ (where $N^m_i$ represents the number of feasible motions for each part $P_i$), it becomes increasingly imperative to develop efficient matrix calculation methods as the number of parts increases and the calculation time becomes exponentially more significant.

According to the definitions of an interference-free matrix, the matrix in each negative axis direction must be identical to the transposed matrix in each positive axis direction.
The constraint degree matrix signifies the constraints existing between the different parts, and its order is immaterial. Hence, we can replicate the upper triangular components with its corresponding lower triangular components.
Similarly, the lower triangular components of contact matrix and constraint degree matrix can be derived from the corresponding upper triangular components.

To calculate the interference-free matrix and constraint degree matrix, we employ a simulation using a CAD model. Specifically, we place the two target parts in their assembled pose and check for interference when the target part is displaced in the axial direction relative to the object coordinate system. 
The target part is displaced by the length of the corresponding side of the bounding box that encompasses the two target parts unless interference is detected.
Additionally, we conduct a displacement simulation to generate the constraint degree matrix by altering the pose of the target part according to the specified maximum clearance distance for the target product.

\subsection{Constrained Many-objective Optimization}
The figure depicted in \figref{nsga3} presents a proposed NSGA-III~\cite{NSGAIII}-inspired MaOGA that displays a high level of performance in MaOPs with four or more objectives.
The gene representation is initially arranged according to the user-defined number of genes (Step 1).
To increase the number of effective initial solutions, the CCG is employed.
The evaluation process for multiple objective functions is then carried out (Step 2).
To incorporate the principles of NSGA-III, non-dominated sorting (Step 3) and reference-line-based gene selection (Step 4) are applied.
The algorithm proceeds to generate the next chromosome (Step 6) and apply genetic operations (Step 7) unless the termination condition is met (Step 5).
 
The genetic encoding of information onto chromosomes and the subsequent genetic operations rely on existing methods, the efficacy of which has been demonstrated in the enhancement of ASP optimization~\cite{Tariki2021}.
These operations are subsequently applied, and the process returns to Step 2 and repeats until the specified termination condition is satisfied.
 
The four objective functions designed in this study are calculated from the information extracted from the 3D CAD model.
This study considers a minimization problem that evaluates all objective functions equally, with 0 being the optimal evaluation value and 1 being the worst.

\subsubsection{Constraints Based on Feasibility and Stability}
If the sequence is either infeasible or unstable, the disassembly operation cannot be executed by either human or robot.
Conversely, if the sequence satisfies all constraints (available), then only those sequences that satisfy the constraints will be evaluated based on other objective functions.

The feasibility of the sequence is determined by checking both the order and motion feasibility.
The order is considered order-feasible if it adheres to the specified conditions regarding the interference-free matrices $\bm{X}^{if}_j~(j=1,\ldots,6)$:
\begin{equation}\forlab{nif}
   \sum_{k=2}^{N^p} \Biggl(\prod_{i=1}^{k-1} \sum_{j=1}^{6} X^{if}_j(P_{O_i},P_{O_k}) > 0 ~?~1~:~0\Biggl) = {N^p} - 1.
\end{equation}

The order is considered motion-feasible if one or more collision-free and IK-solvable contacts and trajectories are found for all disassemblies needed to complete the sequence.
Hence, the target sequence is determined as motion-feasible when the following conditions regarding the motion-feasible matrices $\bm{X}^{mf}_i~(i=1,\ldots,N^p)$ are fulfilled.
\begin{equation}\forlab{nmf}
   \sum_{k=2}^{N^p} \Biggl(\prod_{i=1}^{k-1} \sum_{j=1}^{N^m_{pok}} X^{mf}_{pok}(\mathcal{\bm{M}}^{pok}_j,P_{O_i}) > 0 ~?~1~:~0\Biggl) = {N^p}.
\end{equation}
where $X^{mf}_{pok}(\mathcal{M}^{pok}_j,P_{O_i})$ represents the motion feasibility matrix between $i$-th part $P_{O_i}$ and the $j$-th set of feasible motion $\mathcal{\bm{M}}^{pok}_j$ for the target $k$-th part $P_{O_k}$ when $pok$ represents ${P_{O_k}}$.
A set of feasible motions $\mathcal{\bm{M}}^i_j$ comprises feasible contact $\mathcal{\bm{C}}^i_j$, feasible trajectories $\mathcal{\bm{T}}^i_j$, and feasible object placement $\mathcal{\bm{P}}^i_j$.
The number of feasible motions generated for each part is represented by $N^m_i~(i=1,\ldots,N^p)$.

The evaluation of stability involves the assessment of two distinct criteria. The first criterion pertains to whether the parts remaining after removal of the target part can maintain an upright posture in the workplace (upright condition).
The second criterion involves determining whether all parts are interconnected (connection condition).
The upright condition (static stability) can be easily assessed using a method established in the fields of optimal 3D fabrication~\cite{Prevost2013} and balance control of humanoid robots~\cite{Harada2003}.
If flexible fixtures are available to hold various poses of the parts, the upright condition can be disregarded, as it will always be satisfied.
In this study, we employ an array of multiple soft jigs~\cite{Kiyokawa2021jig} as a solution to address this issue.
The connection condition can be easily examined by analyzing the constituent elements.
The sequence is considered stable when the following conditions are fulfilled:
\begin{equation}\forlab{nnc}
   \sum_{k=2}^{N^p} \Biggl(\sum_{i=1}^{k-1} X^{ct}(P_{O_i},P_{O_k}) \not = 0~?~1~:~0\Biggl) = {N^p} - 1.
\end{equation}
The symbol $\bm{X}^{ct}$ denotes a contact matrix, where a value of one indicates the presence of a nonzero value in the constraint degree matrix $\bm{X}^{cs}$.

\subsubsection{Initialization of Chromosomes}
To increase the number of high-quality initial solutions generated, we employ stability-based chromosome initialization utilizing the CCG depicted in~\figref{graph}~(a).
The graph is automatically generated using the following procedure:
\begin{enumerate}
    \item The parts are classified as either bolts or screws (fixing parts, represented by box-shaped nodes) or other parts (non-fixing parts, represented by circle-shaped nodes) based on the task labels extracted through model structure analysis.
    \item Edges are generated between each node by analyzing the contact matrix, connecting nodes that are in contact with each other.
    \item The edges connecting to the fixing part nodes are categorized and assigned as connection edges (red-colored edges) and other edges (black-colored edges).
\end{enumerate}
The numbers in the nodes correspond to the part ids of the disassembly target product.

The following procedure is performed based on the generated CCGs:
\begin{enumerate}
   \item The base-labeled part or the largest part is designated as the root node.
   \item The distance (minimum number of edges) from the root node to each node is calculated.
   \item A node is randomly selected from the set of nodes at the maximum distance.
   \item If the selected part is a fixing part, it is placed at the beginning of the sequence. If it is a non-fixing part, a neighboring part connecting the selected part to another part is randomly selected and placed at the beginning of the sequence. If there are no fixing parts, the selected non-fixing part is placed at the beginning of the sequence.
   \item Steps 2 to 4 are repeated until only the root node remains. The part displaying the root node is then placed at the end of the disassembly sequence, resulting in the end of the generation process.
\end{enumerate}
\figref{graph}~(b) shows snapshots of an example of the disassembly procedure.
The absence of isolated nodes not connected to any edge indicates that every disassembly can be regarded as stable.

The essential prerequisite for attaining an interference-free sequence when dealing with a fully constrained part that impedes motion in all 12 directions is the prioritized removal of the fixing part. 
Therefore, utilizing CCGI is more advantageous for generating interference-free initial solutions than relying on random initialization.
In other words, the use of CCGI is more likely to result in the generation of an available sequence.

Subsequently, the optimization process commences with the initial solutions.
Throughout the optimization procedure, feasible and stable solutions also improve the evaluation of the objective functions.
To ensure uniformly evaluated multiple objectives, the objective values must be normalized, enabling meaningful distance metric computations in the objective space.
Experimental results from a previous study by Blank~\etal~\cite{Blank2019} indicated that normalization affects the performance of evolutionary multi-objective optimization (EMO) algorithms.
Thus, this study normalizes the four objectives to values ranging from zero to one, as described in the following sections.

\subsubsection{Difficulty}
Among the various difficulty definitions~\cite{Kiyokawa2023RCIM}, we propose a specific definition for the objective function that corresponds to the constraint state transition difficulty~\cite{Yoshikawa1991}, which is a type of order difficulty. This can be expressed as follows:
\begin{equation}\forlab{fd}
 f_d
  \coloneqq
 \left\{
  \begin{array}{ll}
     H / 12({N^p}-1)  &  \text{if $\bm{O}$ is available} \\
     1      &  \text{otherwise}
    \end{array}
 \right..
\end{equation}
where $H$ denotes the maximum level of constraint state transition difficulty associated with the disassembly of each individual part.
\begin{equation}\forlab{hs}
   H
   \coloneqq
   \max_{k \in \{2,3,\ldots,{N^p}\}} \sum_{i=1}^{k-1} X^{cs}(P_{O_i},P_{O_k})~< 12({N^p}-1).
\end{equation}
$\sum_{i=1}^{k-1} X^{cs}(P_{O_i},P_{O_k})$ represents the $k$-th part $P_{O_k}$ and its undisassembled parts $P_{O_1}, P_{O_2}, \ldots, P_{O_{k-1}}$. In accordance with the established definition, the maximum constraint degree between the two parts is 12, consequently, each element of the constraint degree matrix $\bm{X}^{cs}$ is calculated as
\begin{equation}\forlab{cik}
   X^{cs}(P_i,P_k)
   =
   12 - \sum_{j=1}^{12} X^{cf}_j(P_i,P_k)~\in \{0,\ldots,11\}.
\end{equation}
The matrix $\bm{X}^{cf}_j~(j=1,\ldots,12)$ represents the constraint-free matrix.

\subsubsection{Efficiency}
The task labels are utilized to maximize the efficiency of the sequence of tasks by minimizing the number of task changes and the distance between the center of mass of the target parts.
\begin{equation}\forlab{fe}
 f_e
  \coloneqq
 \left\{
  \begin{array}{ll}
     [N^{tc}/({N^p}-1) \\~~+ D/ ({N^p} \times D_{\rm{max}})] / 2 &  \text{if $\bm{O}$ is available} \\
     1      &  \text{otherwise}
    \end{array}
 \right..
\end{equation}
The number of task changes $N^{tc}$ can be determined by analyzing the task labels $T_{O_i}$ of each part $i=1,\ldots,{N^p}$.
\begin{equation}\forlab{ntc}
    N^{tc}
    =
    \sum_{k=2}^{N^p} \left[(T_{O_k} = T_{O_{k-1}})~?~1~:~0\right].
\end{equation}
The total moving distance $D$ can be determined by measuring the distance $d_{P_i,P_j}$ between each part.
\begin{equation}\forlab{d}
   D
   =
   \sum_{k=2}^{N^p} d_{P_{O_k},P_{O_{k-1}}}.
\end{equation}
The maximum distance between any two parts is denoted by $D_{\rm{max}}$.

\subsubsection{Prioritization}
The objective function for prioritization is defined as follows:
\begin{equation}\forlab{fp}
 f_p
  \coloneqq
 \left\{
  \begin{array}{ll}
     1 - R / \prod_{l={N^p}-N^{pp}}^{N^p} l &  \text{if $\bm{O}$ is available} \\
     1      &  \text{otherwise}
    \end{array}
 \right..
\end{equation}
$N^{pp}$ denotes the number of prioritized parts. The degree of prioritization $R$ is determined by the positions of priority parts.
\begin{equation}\forlab{r}
   R = \sum_{m=1}^{N^{pp}} O^p_m.
\end{equation}
$O^p_m$ represents the ordinal position of the $m$-th priority part.

\subsubsection{Allocability}
Allocability is based on the sequential position of the manually labeled parts to be disassembled.
\begin{equation}\forlab{fa}
 f_a
  \coloneqq
 \left\{
  \begin{array}{ll}
     |O^m_r - O^m_l| / ({N^p}-1) &  \text{if $\bm{O}$ is available} \\
     1      &  \text{otherwise}
    \end{array}
 \right..
\end{equation}
$O^m_l$ and $O^m_r$ indicate the latest and earliest ordinal positions of manually disassembled parts, respectively.

\section{Experiments of Robotic DSP}
\begin{figure}[tb]
  \centering
  \includegraphics[width=0.9\linewidth]{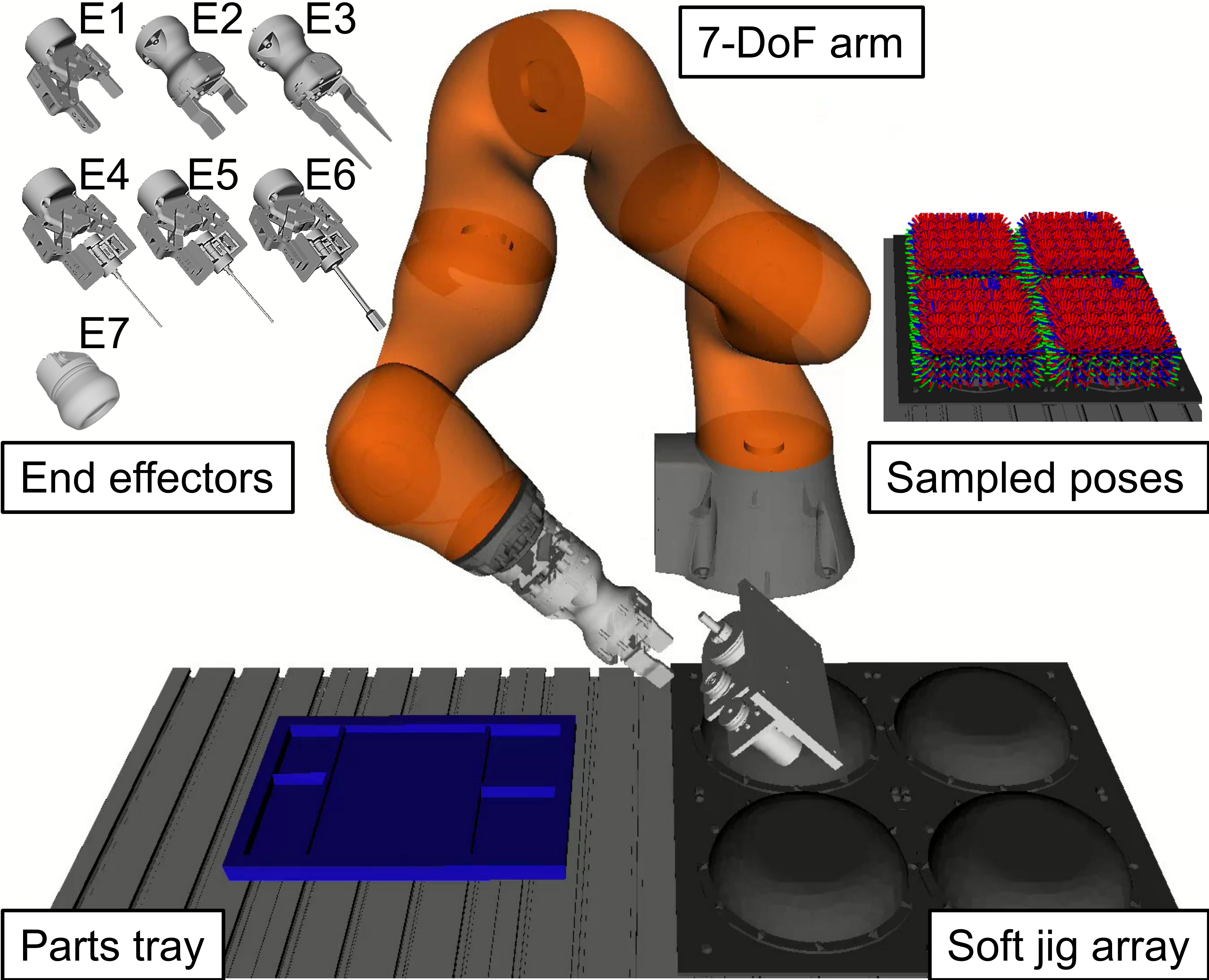}
  \caption{\small{Robotic disassembly setup for simulation experiments.}}
  \figlab{system}
\end{figure}
\begin{figure}[tb]
  \centering
  \includegraphics[width=\linewidth]{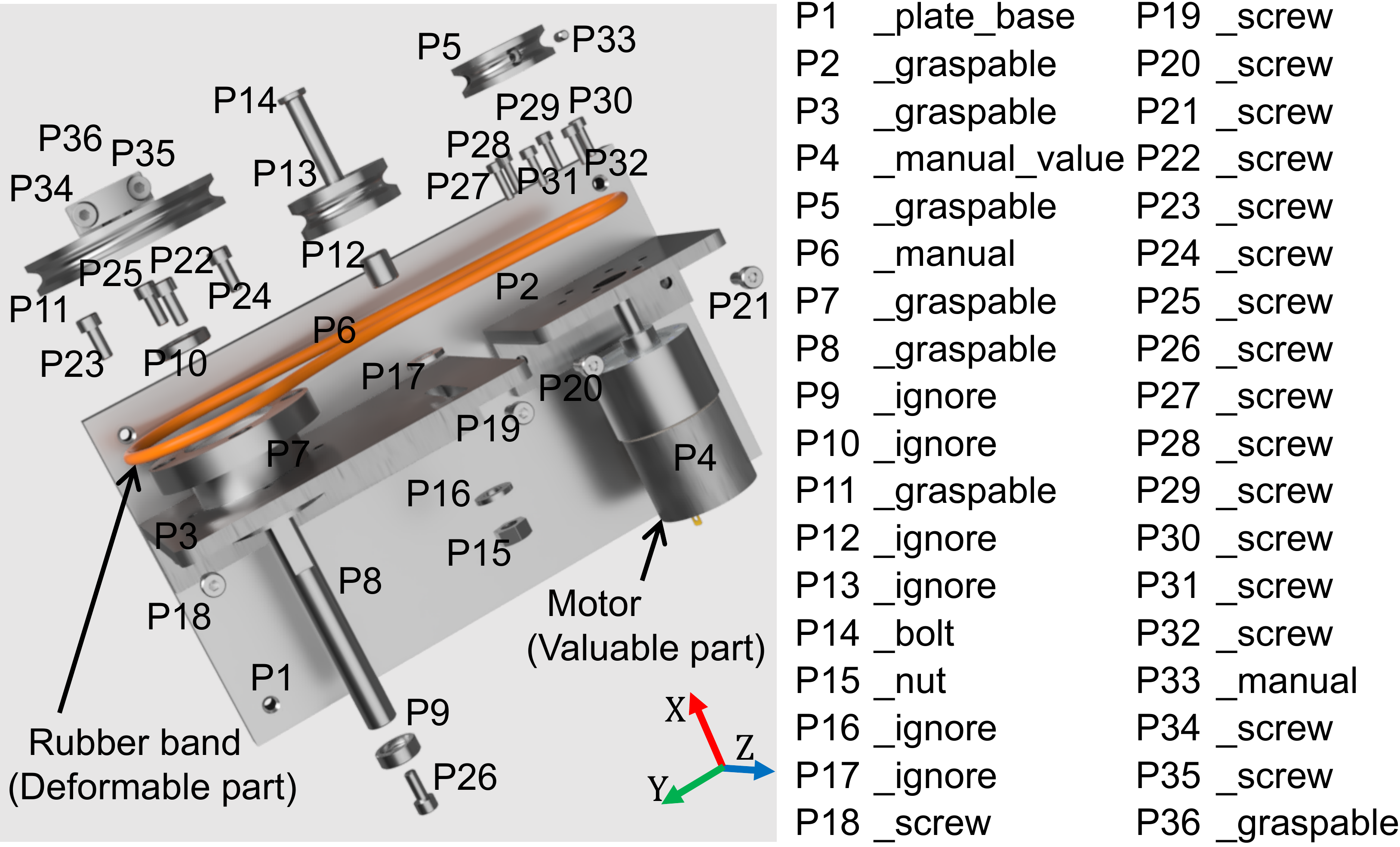}
  \caption{\small{Model appearance and assigned parts labels.}}
  \figlab{obj}
\end{figure}
\begin{figure}[tb]
  \centering
  \begin{minipage}[tb]{0.49\linewidth}
      \centering
      \includegraphics[width=\linewidth]{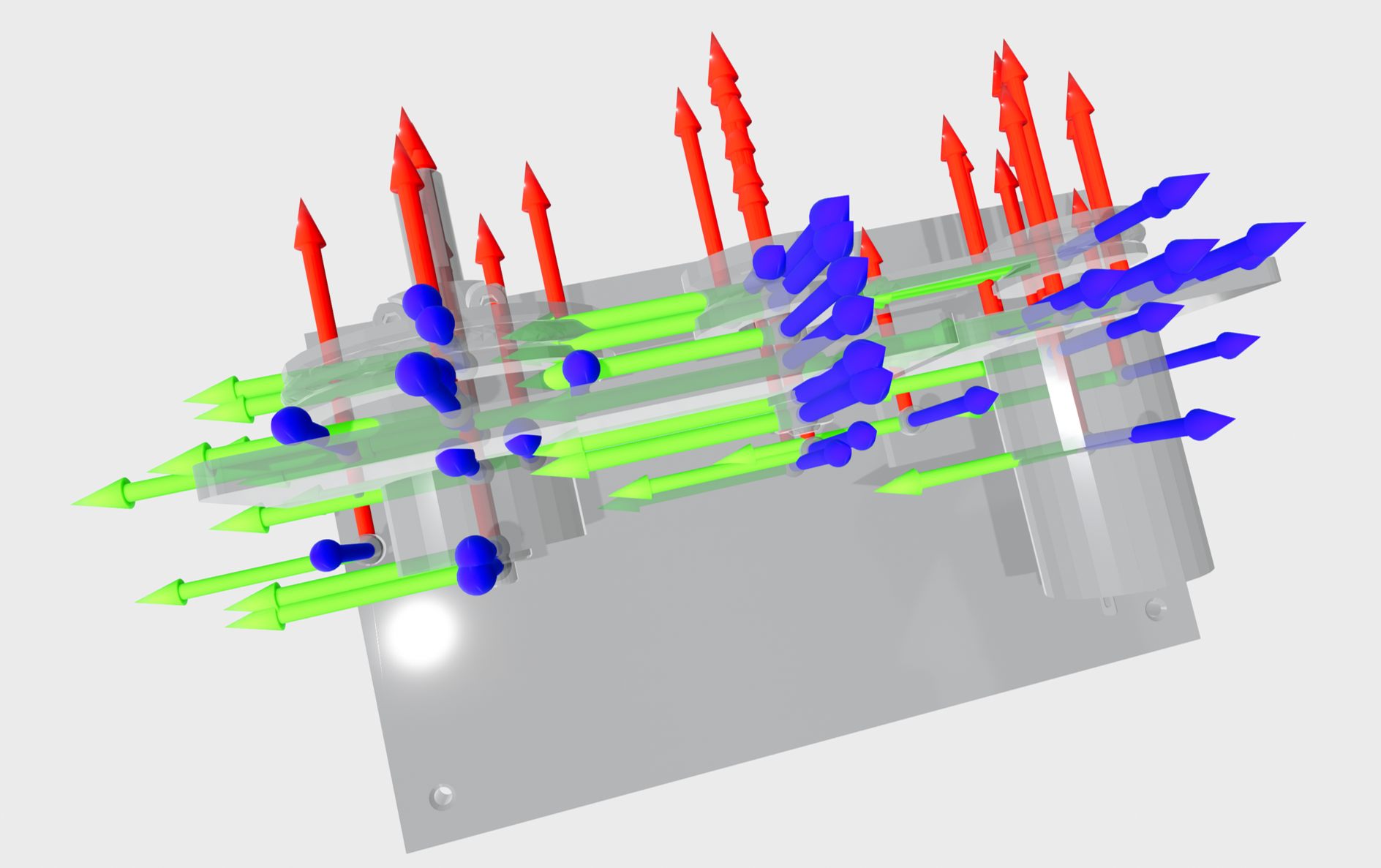}
      \subcaption{}
  \end{minipage}
  \begin{minipage}[tb]{0.49\linewidth}
      \centering
      \includegraphics[width=\linewidth]{./figure/check.pdf}
      \subcaption{}
  \end{minipage}
  \caption{\small{Results of structure analysis and interference check for the belt drive unit. (a) Center of mass and pose. (b) Example of interference.}}
  \figlab{modules}
\end{figure}
\subsection{Overview}
Our experiments verified the efficacy of the proposed method in terms of structure analysis, matrix generation, and DSP using the robotic disassembly setup illustrated in~\figref{system}.
The objective of our experiments was to evaluate the performance of the proposed method on a belt drive unit used in an assembly challenge~\cite{WRS2018}.
\figref{obj} depicts the appearance of the CAD model and shows the labels assigned for our experiments.

The product disassembly system incorporated a seven-degree-of-freedom (DoF) arm and various eefs.
Three different types of two-finger parallel grippers were utilized: the Robotiq 2F-85 (E1), the Robotiq HandE (E2), and the Robotiq HandE with longer fingers (E3).
In order to enable the robot arm to screw bolts and screws using a two-finger parallel gripper, Hu~\etal~\cite{Hu2022} developed a mechanical screwing tool.
The system utilized three configurations of the screwing tool with different tool tip parts: m3 hex wrench (E4), m4 hex wrench (E5), and m6 socket wrench (E6). The suction gripper employed was the CONVUM SGB30 (E7), also known as the balloon hand, which is well suited for handling a wide range of workpiece shapes and sizes and allows for easy handling of uneven, heavy, porous, and other workpieces.

The process of structural analysis involves identifying the labels assigned to various parts.
The belt drive unit comprises the lables of screw, bolt, nut, plate, graspable, manual, value, base, and ignore.
The P4 motor was regarded as a valuable part, hence it possesses a value label in addition to its manual label, owing to its delicate disassembly process.
The rubber belt P6 and hexagon socket set screw P35 were also assigned a manual label because of their difficulty in robotic disassembly.
The spacers P9, P10, and P12, pulley P13, and washers P16 and P17 were assigned the ignore label, as they will naturally come off during the disassembly process of other parts.

The allocation of the seven types of eefs was determined according to the task label and shape features for 27 of the 36 parts. These 27 parts, namely P1, P2, P3, P5, P7, P8, P11, P14, P15, P18, P19, P20, P21, P22, P23, P24, P25, P26, P27, P28, P29, P30, P31, P32, P34, P35, and P36 did not have a manual or ignore label.
Therefore, for these parts, we have assigned E7, E2, E2, E2, E2, E1, E2, E5, E6, E5, E5, E5, E5, E5, E5, E5, E4, E4, E4, E4, E4, E4, E5, E5, and E3.

The generation of sequences requires several parameters $\bm{P}^{ga}$, including the number of chromosomes, crossover rate, mutation rate, cut-and-paste rate, and break-and-join rate, which are identical to those utilized in~\cite{Kiyokawa2021seq}.
In our methodology, we set the number of generation updates to 500, and the number of iterations to 10.
To create robot motions, we utilize a contact planning software\footnote{https://github.com/Osaka-University-Harada-Laboratory/wros} that uses an object-geometry-based approach to potential contact generation, as described in~\cite{Wan2021}.
The robot arm trajectory planner relies on the RRT-connect algorithm~\cite{Kuffner2000}, which was implemented in MoveIt! motion planning framework\footnote{https://moveit.ros.org/} of Robot Operating System (ROS), as well as IKFast~\cite{diankov_thesis} for solving the kinematics.
In a work environment equipped with four soft jigs arranged on the workspace, as depicted in the upper right corner of~\figref{system}, during the processing of RobotSimulation($\bm{M}$) in~\algoref{dsp}, the target assembled parts were positioned and oriented in the softjig array.
Thereafter, a trajectory was explored to identify potential collision-free contact points and efficiently executable trajectories for the target product at these specific positions and orientations.

\subsection{Results of CAD-Informed Matrices Generations}
In order to generate multiple matrices for determining the constraints and calculating the evaluation values in the optimization, we initially conducted a thorough analysis of the 3D CAD models.
Our analysis successfully extracted all part labels with a high degree of accuracy.
The center of mass and pose parameters obtained from the STEP model were accurately extracted with 100\% accuracy, as illustrated in~\figref{modules}~(a).
The interference check between parts for matrix generation is shown in~\figref{modules}~(b), where the red highlighted area indicates the interfered volume between the base plate part P1 and L-shaped plate part P2.

We assessed the performance of the automatic matrix extraction based on accuracies.
The constraint degree matrix contains positive integers, and the calculation was regarded as successful when a positive integer was correctly determined to match the manually annotated value.

The generation accuracies of the interference-free and contact matrices were 98.9\% and 96.8\%, respectively.
The success rates of the constraint degree and constraint-free matrices were 90.2\% and 92.9\%, respectively, which were not low.
The interference checks based on the displacement simulation can sometimes fail due to the limitations of the Boolean operation performance. The Boolean operations between curved surfaces can be challenging and may result in errors. In the future, it may be necessary to consider a method for directly estimating the degree of constraint based on the shape.

\subsection{Optimizing Sequences}
\begin{figure}[tb]
  \centering
  \includegraphics[width=\linewidth]{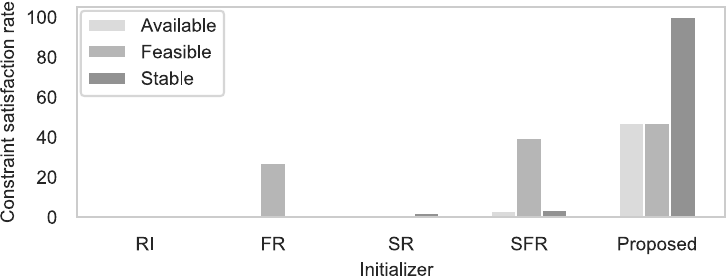}
  \caption{\small{Success rates of finding constraint-satisfied sequences in 1000-trial initialization [\%]. RI, FR, SR, and SFR are comparative methods that show random initialization, feasibility-based rearrangement, stability-based rearrangement, and stability-and-feasibility-based rearrangement methods, respectively.}}
  \figlab{init}
\end{figure}
\begin{figure*}[tb]
  \centering
  \includegraphics[width=\linewidth]{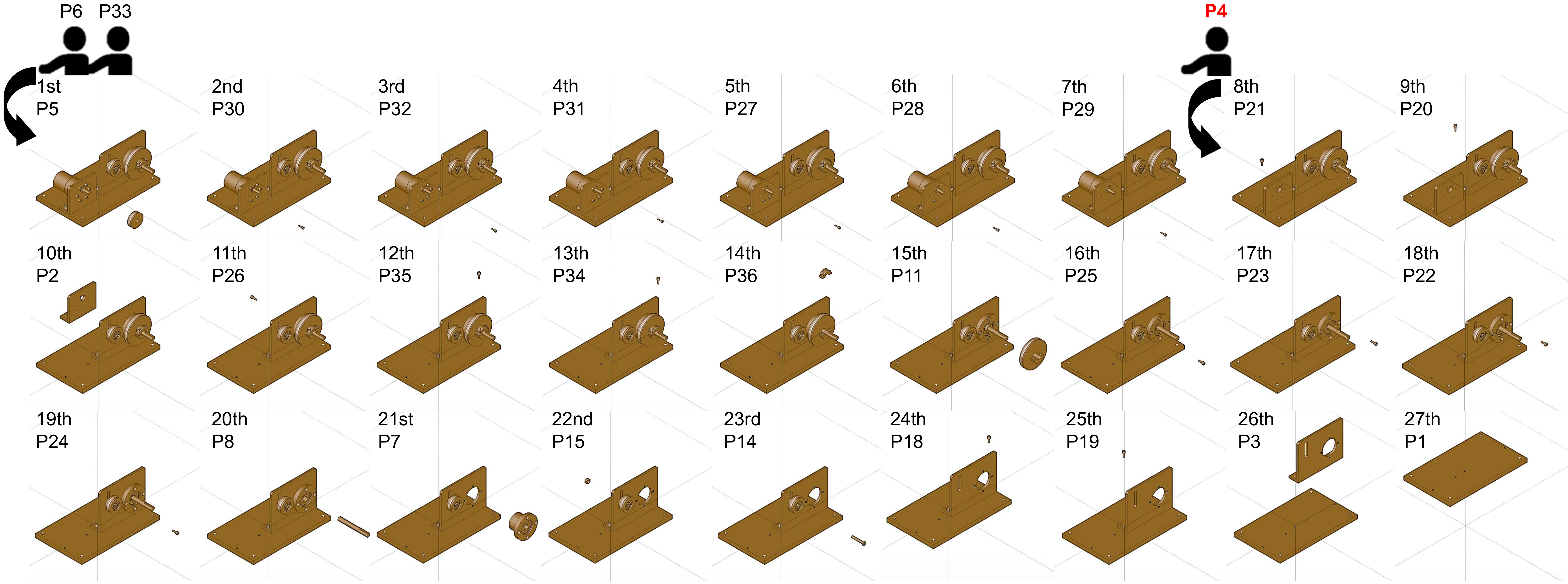}
  \caption{\small{A sequence determined after the optimization. The disassemblies shown in the snapshots are apparently order feasible (interference-free) and stable.}}
  \figlab{seq}
\end{figure*}
\begin{figure}[tb]
  \centering
  \begin{minipage}[tb]{\linewidth}
      \centering
      \includegraphics[width=\linewidth]{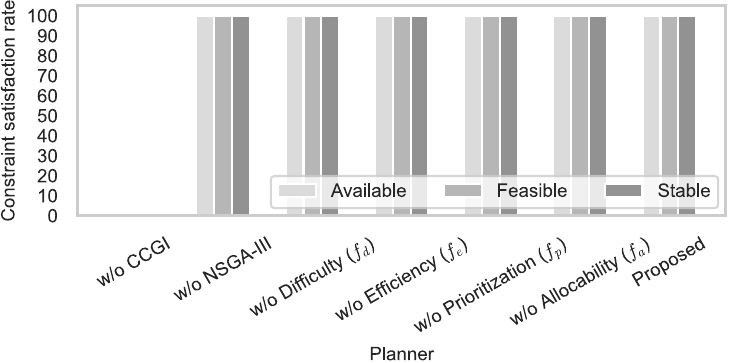}
      \subcaption{}
  \end{minipage}
  \begin{minipage}[tb]{\linewidth}
      \centering
      \includegraphics[width=\linewidth]{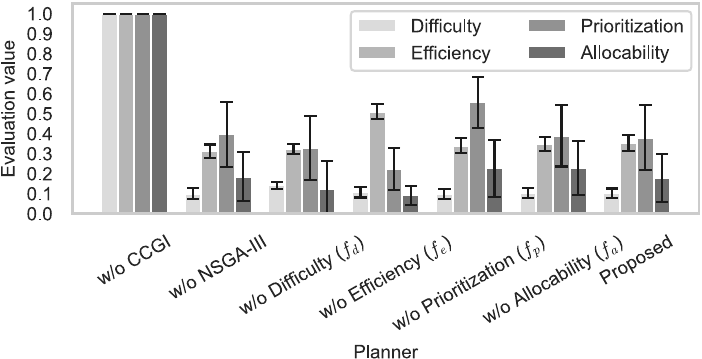}
      \subcaption{}
  \end{minipage}
  \caption{\small{The many-objective optimization results of 500-generation and 10-iteration optimization. (a) Percentages of constraint-satisfied solutions [\%]. (b) Mean $\pm$ standard deviation of the evaluation values.}}
  \figlab{opt-obj}
\end{figure}
\begin{figure*}[tb]
  \centering
  \begin{minipage}[tb]{\linewidth}
      \centering
      \includegraphics[width=\linewidth]{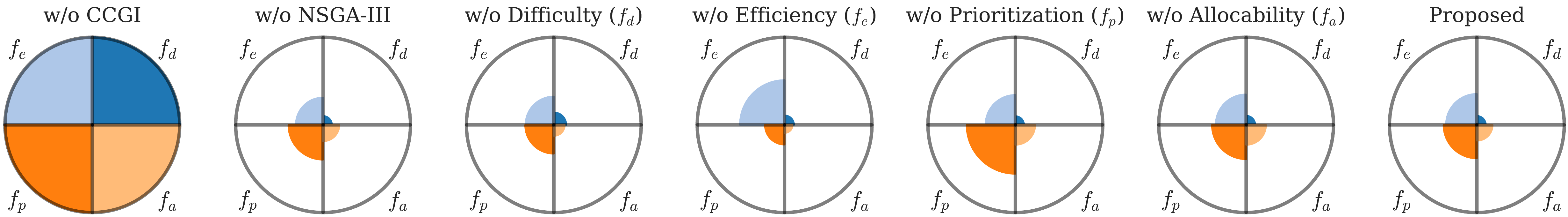}
      \subcaption{}
  \end{minipage}
  \begin{minipage}[tb]{\linewidth}
      \centering
      \includegraphics[width=\linewidth]{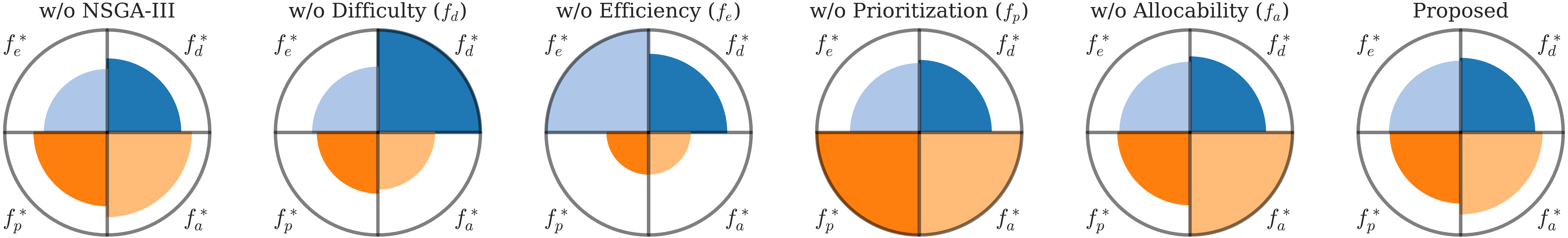}
      \subcaption{}
  \end{minipage}
  \begin{minipage}[tb]{\linewidth}
      \centering
      \includegraphics[width=\linewidth]{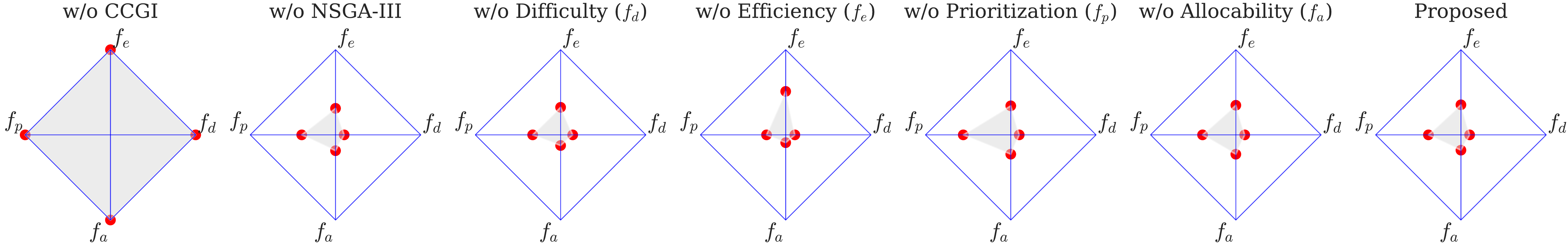}
      \subcaption{}
  \end{minipage}
  \begin{minipage}[tb]{\linewidth}
      \centering
      \includegraphics[width=\linewidth]{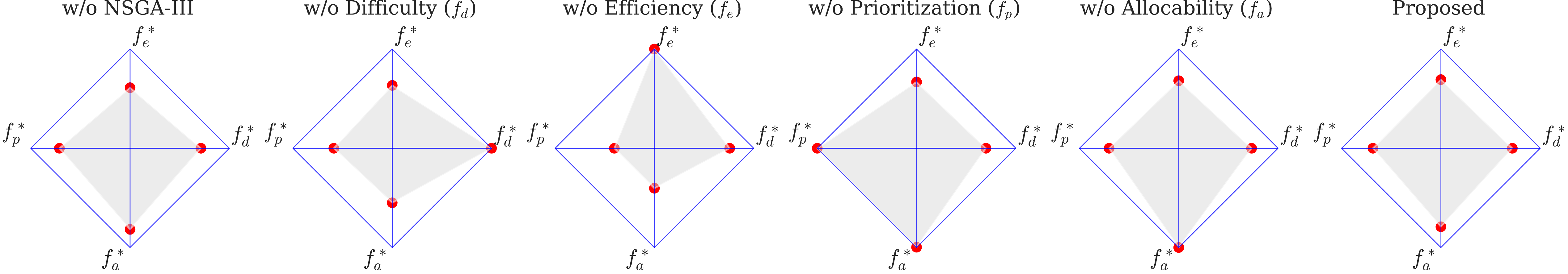}
      \subcaption{}
  \end{minipage}
  \caption{\small{Visualization of the mean evaluation values over the generation updates for all iterations. (a) Petal chart (original). (b) Petal chart (relative evaluation). (c) Radar chart (original). (d) Radar chart (relative evaluation). For the petal chart, the smaller each petal, the better is the evaluation value. Regarding the radar chart, the smaller the square area, the better the evaluation value. The graphs for the relative evaluation were min-max normalized so that the maximum value of each objective function in all methods except w/o CCGI was scaled to 1.0.}}
  \figlab{charts}
\end{figure*}
\begin{figure}[tb]
  \centering
  \begin{minipage}[tb]{\linewidth}
      \centering
      \includegraphics[width=\linewidth]{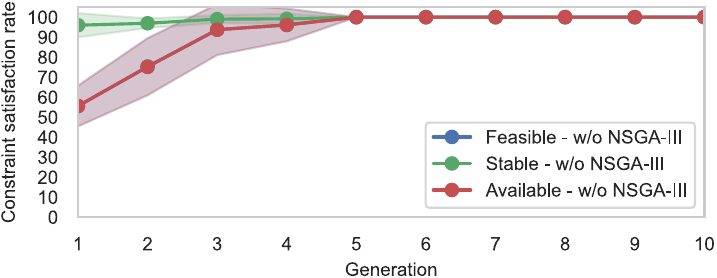}
      \subcaption{}
  \end{minipage}
  \begin{minipage}[tb]{\linewidth}
      \centering
      \includegraphics[width=\linewidth]{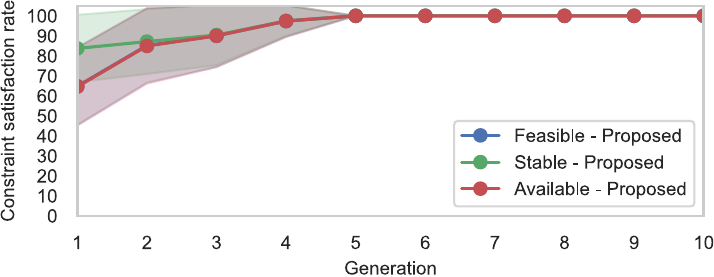}
      \subcaption{}
  \end{minipage}
  \caption{\small{Changes in the mean values (over 10 iterations) of the constraint satisfaction rates [\%] during 500 generation updates in the first iteration. (a) w/o NSGA-III. (b) Proposed. After the fifth generation, all constraint satisfaction rates were 100\%. The translucent bands around the lines illustrate the standard deviations (over 10 iterations).}}
  \figlab{conv-rates}
\end{figure}
\begin{figure}[tb]
  \centering
  \begin{minipage}[tb]{\linewidth}
      \centering
      \includegraphics[width=\linewidth]{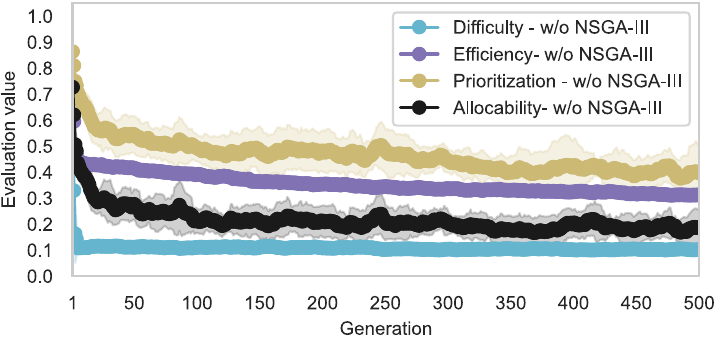}
      \subcaption{}
  \end{minipage}
  \begin{minipage}[tb]{\linewidth}
      \centering
      \includegraphics[width=\linewidth]{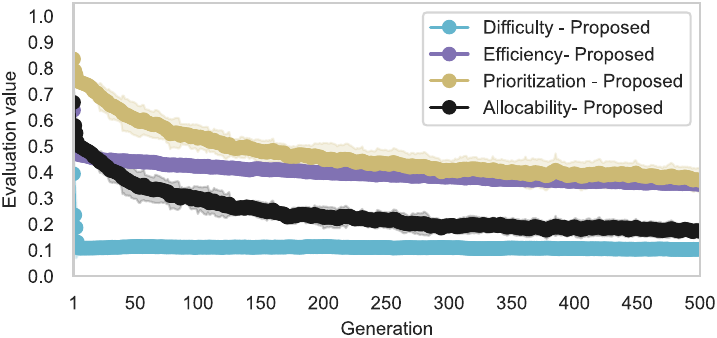}
      \subcaption{}
  \end{minipage}
  \caption{\small{Changes in the mean evaluation values (over 10 iterations) during 500 generation updates in the first iteration. (a) w/o NSGA-III. (b) Proposed. The translucent bands around the lines illustrate the standard deviations (over 10 iterations).}}
  \figlab{conv-evals}
\end{figure}
\begin{figure}[tb]
  \centering
  \includegraphics[width=\linewidth]{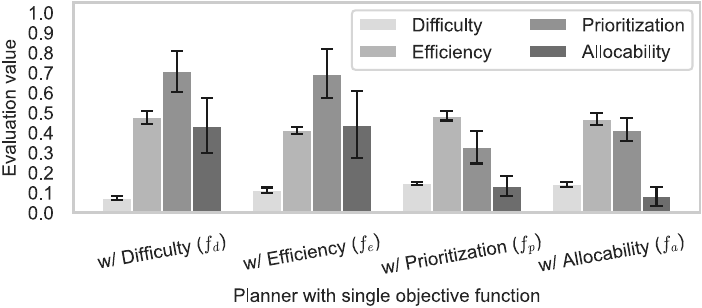}
  \caption{\small{Final evaluation values of single-objective optimization. The mean $\pm$ standard deviation of the evaluation values ranged from 0 to 1 in the 500-generation and 10-iteration optimizations [\%].}}
  \figlab{opt-single-obj}
\end{figure}
\begin{figure*}[tb]
  \centering
  \includegraphics[width=\linewidth]{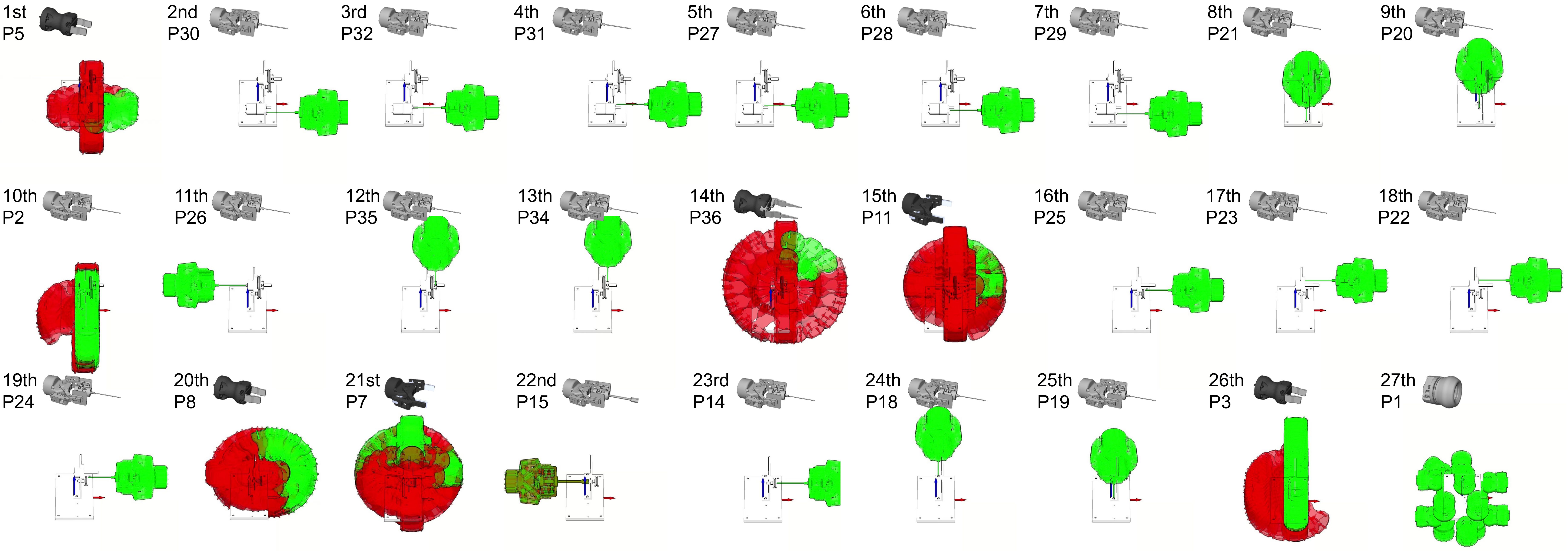}
  \caption{\small{The generated feasible contacts $\bm{\mathcal{C}}$ for each target part by the eefs. The green eefs show collision-free contacts that do not interfere with the target object. The red circles show the possible failed contacts owing to collisions.}}
  \figlab{contact}
\end{figure*}
\begin{figure*}[tb]
  \centering
  \includegraphics[width=\linewidth]{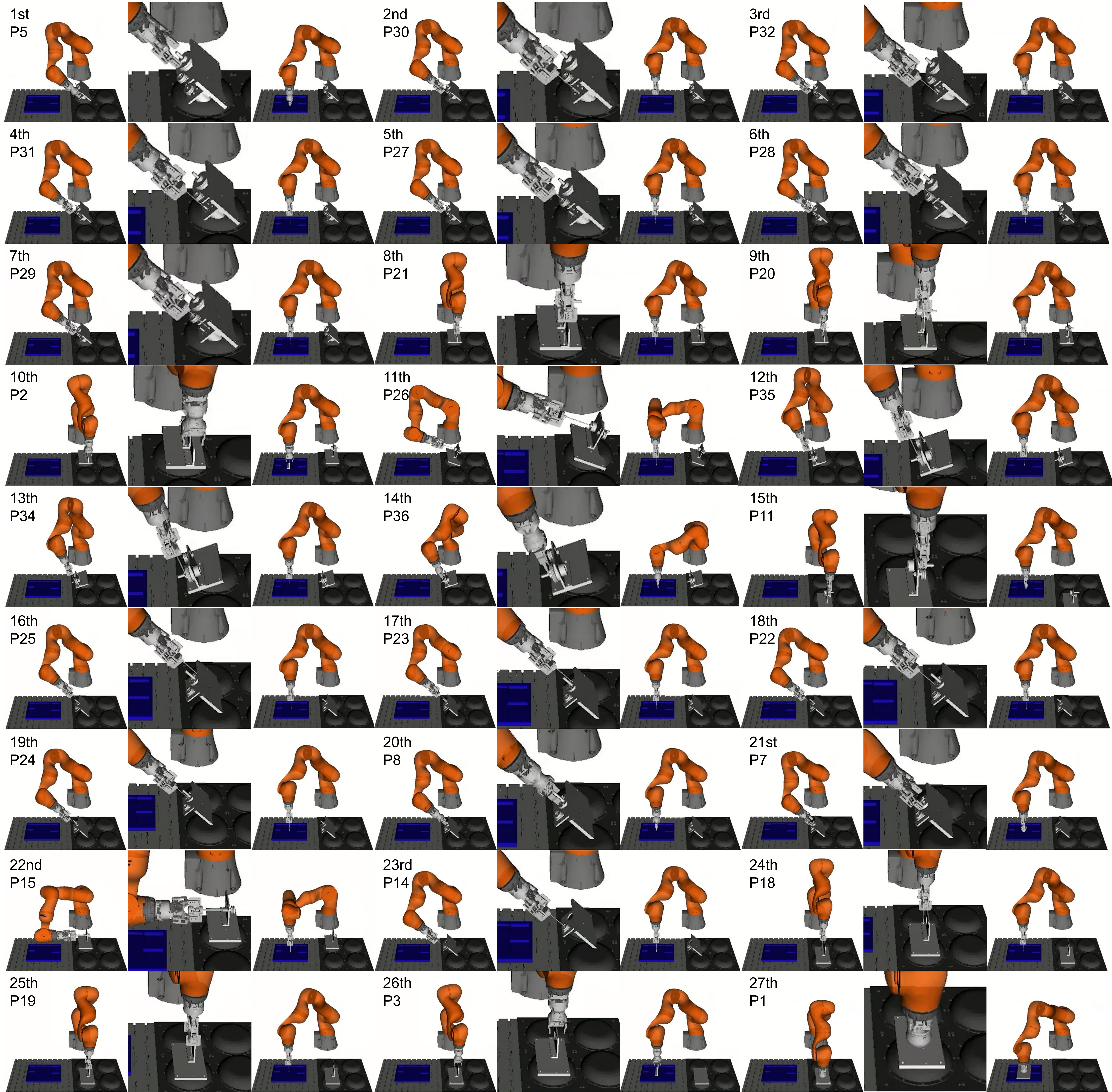}
  \caption{\small{The generated feasible (collision-free and IK-solvable) motions include $\bm{\mathcal{\hat{C}}}$, $\bm{\mathcal{\hat{T}}}$, and $\bm{\mathcal{\hat{P}}}$. The three pictures for each disassembly show the arm in a pose at the post-contact position, the zoom of the post-contact pose, and the placement pose.}}
  \figlab{motion}
\end{figure*}
\subsubsection{Performance of Choromosome Initialization}
We undertook a comparative analysis of chromosome initialization methods.
To this end, we devised three methods for initializing chromosomes: random initialization (RI), repeated sequence changes to minimize interference (FR)~\cite{Tariki2021}, and repeated sequence changes to maximize stability (SFR).
\figref{init} shows the result comparisons.
The bars depict the mean values of the feasible, stable, and available (feasible and stable) rates [\%] for the 1000-trial initialization.
Although the rates of feasible solutions with FR and SFR were 27.1\% and 39.6\%, respectively, the rates of stable solutions were 0.2\% and 3.6\% for FR and SFR resulting in the available rates were 0.1\% and 3.3\%, respectively.

Nevertheless, the feasibility, stability, and availability rates for the proposed method were 47.3\%, 100\%, and 47.3\%, respectively.
The results indicate that the proposed method can generate a feasible and stable sequence when compared to other methods.

\subsubsection{Performance of Optimization}
The generated sequence $\bm{\hat{O}}$ composed of elements 1, 3, 19, 18, 14, 15, 7, 8, 24, 22, 23, 25, 11, 36, 34, 35, 26, 2, 20, 21, 4, 29, 28, 27, 31, 32, 30, 5, 33, 6 attained the lowest evaluation value among all evaluated sequences.
This sequence includes parts with manual labels, but excludes those with the ignore label.
This sequence is a feasible and stable (available) solution that adheres to all imposed constraints.

\figref{seq} shows the generated sequence.
As can be observed in the figure, the unconstrained parts P6 and P33 with manual labels, are situated at the commencement of the sequence.
In addition, motor part P4 with manual and value labels, is disassembled at the earliest possible timing following the elimination of all constraining parts, namely P5, P30, P32, P31, P27, P28, and P29.
It is worth noting that other arrangements also fulfill order feasibility, motion feasibility, and stability, while simultaneously exhibiting low difficulty and high efficiency, and adhering to label-defined prioritization.
 
\figref{opt-obj}~(a) presents the performance comparison results. We conducted ten iterations of 500 generation updates for the optimization loop.
\figref{opt-obj}~(a) shows the mean values of the available, feasible, and stable rates.
\figref{opt-obj}~(b) shows the mean and standard deviation of the evaluation values for each objective function.
The bars represent the mean values and error bars indicate the standard deviations.
The w/o CCGI method does not utilize CCGI but instead uses an initialization method that considers only feasibility, as previously described in~\cite{Tariki2021}.
The w/o NSGA-III result is based on the NSGA-II-inspired algorithm proposed in~\cite{Kiyokawa2021seq}.
The w/o $f_d$, w/o $f_e$, w/o $f_p$, and w/o $f_a$ denote the optimization results excluding each of the objective functions of \forref{fd}, \forref{fe}, \forref{fp}, and \forref{fa}.
 
The effectiveness of the proposed initialization method in consistently producing available solutions was demonstrated by the lack of solution generation when the CCGI method was not employed, resulting in a 0\% rate of available solution generation.
Comparing the proposed method with the w/o NSGA-III method, both achieved 100\% success rates in generating available solutions.
When evaluating the performance of each method based on the four objective functions, Difficulty, Efficiency, Prioritization, and Allocability, the mean evaluation values of both of them are almost the same.

The visualizations in~\figref{charts} display the mean evaluation values of the final solutions after each optimization iteration.
For the petal chart, smaller petals indicate better evaluation values.
In the case of the radar chart, a smaller square area signifies a better evaluation value.
The graphs (original) show the values calculated using $f_d$, $f_e$, $f_p$, and $f_a$.
The graphs for the relative evaluation values $f_d^*$, $f_e^*$, $f_p^*$, and $f_a^*$ were min-max normalized to scale the maximum value of each objective function in all methods except w/o CCGI to 1.0.
\figref{conv-rates} dipicts the transitions of constraint satisfaction rates for the w/o NSGA-III and proposed methods.
\figref{conv-evals} dipicts the learning curves for the four objective functions evaluated in the cases of using their two methods.

As depicted in the charts presented in~\figref{charts} (a) and (c), the performance of the proposed method and other comparative methods may not exhibit significant disparities.
In fact, the sum of our evaluation values (original) for w/o CCGI, w/o Difficulty, w/o Efficiency, w/o Prioritization, w/o Availability, and Proposed amounted to 4.0, 0.918, 0.934, 1.22, 1.07, and 1.01, respectively.
On the other hand, the standard deviations of the normalized evaluation values were 0.073, 0.180, 0.253, 0.158, 0.129, and 0.0420, respectively.
Notably, the proposed method exhibited the lowest standard deviation. This result indicates that the proposed method was effective in optimizing the system while simultaneously evaluating four objective functions.

As evidenced in~\figref{conv-rates}, although both methods achieved a constraint satisfaction rate of 100\% by the 5-th generation, the proposed method demonstrated a more rapid improvement in these rates up to that point.
Additionally, as depicted in~\figref{conv-evals}, the proposed method achieves a smoother and more consistent learning curve compared to the w/o NSGA-III method, suggesting that the proposed method consistently and stably reduced the evaluation values of the four objective functions throughout the learning process.
These findings suggest that the NSGA-III-inspired algorithm is beneficial for facilitating convergence in the learning process.
In forthcoming studies, we intend to further explore the effectiveness of this feature by applying the proposed method to a variety of other target objects.

\figref{opt-single-obj} shows a comparison of performance using methods that employ a single objective function.
The bars above the titles of w/ Difficulty, w/ Efficiency, w/ Prioritization, and w/ Allocability represent the mean values of the four objective functions when optimizing the solutions under constrains and a single objective function of $f_d$, $f_e$, $f_p$, and $f_a$, respectively.
The error bars indicate standard deviations.
The evaluation value using $f_d$ for w/ Difficulty method is 0.075, which is the lowest values compared to them of other methods.
The difficulty evaluation values of w/ Efficiency, w/ Prioritization, and w/ Allocability methods are 0.112, 0.147, and 0.142, respectively.
As shown in~\figref{opt-single-obj}, the other three evaluation values exhibited the same trend.
The results shown in~\figref{opt-single-obj} demonstrate that the proposed algorithm can effectively perform single-objective optimization, allowing users to choose the objective function that they wish to prioritize.

\subsection{Feasibility of Robotic Disassembly}
\figref{contact} illustrates the feasible contacts by the task-tailored eefs.
The eefs colored in green indicate successful contacts, while those colored in red represent failed contacts resulting from collisions.
We chose the approachable contact from the robot's eef pose among the feasible contacts $\bm{\mathcal{C}}$.

\figref{motion} illustrates the generated feasible (collision-free and IK-solvable) motions for the optimal sequence for robotic disassembly.
\figref{motion} includes snapshots of the generated eef contact $\bm{\mathcal{\hat{C}}}$, arm trajectory $\bm{\mathcal{\hat{T}}}$, and object placement pose $\bm{\mathcal{\hat{P}}}$ fixed on the softjig array.
For each disassembly, the three pictures show the arm in a pose at the post-contact position, the zoom of post-contact pose, and placement pose.
The three pictures for each disassembly show the arm in a pose at the post-contact position, the zoom of post-contact pose, and placement pose.
It is crucial to place the robot in such a position that the target part falls within the arm's movable range.
In some cases, moving the target part to the arm side using a turntable or a flexible fixture placed beneath the object can be effective even for large objects.
Our solution was to use a soft jig, which was previously developed in~\cite{Kiyokawa2021jig,Sakuma2022,Sakuma2023}.
The specific approach may vary based on the configuration of the robot arm, target part, and workspace, utilizing a soft jig array such as ours (\figref{system}), which can fix the target parts in various positions and orientations, enabling the generation of a greater number of trajectory candidates that facilitate the arm approach to the target part.

Although this study did not specifically focus on robot motion planning for real-world tasks,  the findings nonetheless revealed the capability to devise feasible contact for all parts using their corresponding eefs, as well as the corresponding trajectory for all pick-and-place tasks, spanning the entire sequence.

\section{Discussions}
\subsection{Automatic Labeling}
The current approaches to labeling the parts written in the STEP file offer potential for improvement. This study proposes the feasibility of manually inputting explicit labels for object names. However, to minimize the burden on the model designer, it is desirable to develop a method for automatically extracting labels from the geometric information. It is recommended to explore tailored automatic extraction methods for each product, taking into account user convenience.

The PointNet series~\cite{Qi2017,Qi2017++,Qian2022} has demonstrated the potential to classify 3D shape data. Nevertheless, it remains uncertain whether this method can effectively derive task and base labels from the geometric characteristics of parts, as these labels are contingent upon the CAD model, presenting a challenge for automatic labeling.

\subsection{Analysis Time}
The time required for each module in this study was 12.7 seconds, 16400 seconds (4.57 hours), and 1050 seconds (17.5 minutes) for model structure analysis, matrix generation, and sequence exploring, respectively.
The computational time depends on the size of the STEP file and the number of parts, which were 13.8 [MB] and 36 parts, respectively.
The model in the STEP file was represented as a solid model, enabling the calculation of the center of mass. However, when multiple fine curved surfaces, such as fillets, are represented precisely, the cost of calculation becomes excessive when conducting Boolean operations to assess interferences among them.

In our future work, we aim to develop a method that simplifies the shape and reduces the computational costs by eliminating unnecessary parts for the DSP. By utilizing the latest semantic shape representations~\cite{Park2019,Hao2020}, we address the challenge of generating semantic primitive shapes while minimizing the data capacity.

\subsection{Optimization Algorithm}
Several extended NSGA-III algorithms have been proposed.
Among them is U-NSGA-III, which was implemented based on~\cite{UNSGAIII}.
NSGA-III randomly selects parents for mating. It has been demonstrated that tournament selection performs better than random selection.
U-NSGA-III incorporated tournament pressure to achieve unified and improved performance over NSGA-III.

R-NSGA-III is an advanced algorithm that extends NSGA-III.
Further information on its implementation can be found in the work of~\cite{RNSGAIII}.
Occasionally, users are interested in finding a part instead of the entire Pareto-optimal front. First, after analyzing the obtained trade-off solutions using an EMO algorithm, the user may be interested in concentrating on a specific preferred region of the Pareto-optimal front, either to obtain additional solutions in the region of interest or to investigate the nature of solutions in the preferred region.
Second, the user may already have a well-articulated preference among objectives and is directly interested in finding preferred solutions. This paper presents a reference-point-based EMO procedure for achieving both of these purposes.

R-NSGA-III extended the NSGA-III procedure by introducing a new reference point generation method according to
user-supplied aspiration points while using the same genetic operators and survival selection processes.
This approach focused on a specific portion of the Pareto optimal front, making it potentially faster than the original NSGA-III procedure.
The primary objective of their study was to efficiently identify Pareto-optimal solutions that are close to the supplied aspiration points.

Our experimental results indicate that there may be varying levels of heterogeneity in the balance of each objective function depending on the target product to be disassembled.
Therefore, it may bebeneficial to first use a general NSGA-III search to identify these heterigeneities and then use a combination of U-NSGA-III and R-NSGA-III to efficiently optimize them in a more narrow search space. This approach may lead to more efficient optimization results.

\subsection{Real-world Disassembly}
In contrast to the process of assembly, which requires precise and meticulous operations, disassembly can sometimes tolerate minor damage to target products, with the exception of high-value parts that must be carefully disassembled. The generation of robotic disassembly operations in simulations, such as cutting or crushing flexible objects or removing parts without disassembling fasteners, presents a significant challenge. A promising new approach to self-supervised learning in a real-world environment holds potential for acquiring disassembly tasks that involve breaking and damage. Future studies will consider constructing such an approach.

On the other hand, there are situations in which every part must be carefully disassembled without compromising the quality of the materials or breaking them. The precise disassembly operations can be achieved by following the reverse sequence of the assembly operations. In recent years, reinforcement learning approaches have garnered attention for enabling robots to perform contact-rich manipulation tasks in real-world environments, thereby bridging the gap between simulation and reality~\cite{Aguinaco2023}. It may be possible to address the learning problem of disassembly action policy inference model in a similar manner, by following the analogy between assembly and disassembly.

\subsection{HRC-Oriented DSP}
The proposed method for DSP is not limited to the domain of robot motion generation but may also prove effective in teaching sequences to human workers.
By selecting the desired product model using a tablet computer, the results of the automatic DSP can be presented as a guide to efficiently determine the order of the disassembly parts.

This study represents an initial effort to simplify sequencing the semi-automated disassembly operations.
While other criteria for evaluating sequences exist, such as those organized in~\cite{Coronado2022}, this study focused solely on four perspectives: difficulty, efficiency, prioritization, and allocability.
As not all objective functions created based on the provider's motivation can be validated, this study sought a solution for one example design using an EMaO approach.
Kiyokawa~\etal~\cite{Kiyokawa2023RCIM} provide further definitions of the difficulty and complexity.

Previous studies have explored the application of graph representation to determine the necessary operations, tasks, motions, arms, and tools for cooking and furniture assembly sequences using robots~\cite{Takata2022,Sakib2022,Takata2022journal,Wang2023}.
The use of graph-based methods for determining arm and tool availability, calculating efficiency, and determining difficulty levels at different stages represents a promising direction for developing a more general disassembly sequence planner.
If we could design the method to encode the graph into the genes and genetic operations based on the encoded representation, it could potentially be accomplished.

\section{Conclusion}
This study focused on disassembly sequence planning (DSP) that incorporates semi-automated robotic operations.
The proposed robotic DSP method uses an EMaO algorithm, namely NSGA-III-inspired MaOGA that iteratively updates generations and evaluates them with multiple objective functions and constraints.

The results of the disassembly sequence planning for a mechanical product with 36 parts showed that the proposed method can find a Pareto optimal solution oriented towards semi-automated robotic operations.
The algorithm successfully generated a sequence that satisfied the feasibility, stability, and improved conditions in terms of difficulty, efficiency, prioritization, and allocability functions.

Specifically, the use of contact and connection graph (CCG)-based initialization allows for the repeatable generation of a large number of available initial solutions, whereas the algorithm utilized non-dominated sorting and niching with reference lines to encourage steady and stable exploration of the solutions and uniformly lower overall evaluation values.
The final solution featured interference-free, stable, efficient, easy-to-handle, correctly prioritized, and non-redundantly task-assigned order that enables robots collision-free, IK-solvable, and efficient motions in the context of semi-automated robotic disassembly operations.

\section*{Acknowledgement}
This research was carried out at the Panasonic Fundamental Research Collaboration between Osaka University and Panasonic Holdings Co., Ltd.

\bibliographystyle{IEEEtran}
\bibliography{paper.bib}

\end{document}